\documentclass[letterpaper,journal]{IEEEtran}
\usepackage{amsmath,amsfonts}
\usepackage{algorithmic}
\usepackage{algorithm}
\usepackage{array}
\usepackage{textcomp}
\usepackage{stfloats}
\usepackage{url}
\usepackage{verbatim}
\usepackage{graphicx}
\usepackage{cite}
\usepackage{multirow}
\usepackage{enumitem}
\usepackage{lineno}
\usepackage[breaklinks=true,colorlinks,linkcolor=blue,citecolor=blue,urlcolor=black]{hyperref}

\usepackage{flushend}
\usepackage{bbding}
\usepackage{xspace}
\usepackage{booktabs}
\usepackage{multirow}
\usepackage{amsmath,amssymb}
\newcommand{\sexyname}{PGFA\xspace}
\newcommand{\sexynameplus}{PGFA}
\usepackage{xcolor}
\usepackage{colortbl}
\definecolor{mygray}{rgb}{0.9,0.9,0.9}
\def\mL{{\mathcal L}}

\DeclareMathAlphabet\mathbfcal{OMS}{cmsy}{b}{n}
\def\bmD{{\mathbfcal D}}

\def\0{{\bf 0}}
\def\1{{\bf 1}}

\def\bc{{\bf c}}

\def\bp{{\bf p}}

\def\bv{{\bf v}}
\def\bw{{\bf w}}
\def\bx{{\bf x}}

\def\bz{{\bf z}}

\def\mmN{{\mathbb N}}
\def\mmZ{{\mathbb Z}}

\def\mmS{{\mathbb S}}

\def\bx{{\bf x}}

\def\bw{{\bf w}}

\def\bp{{\bf p}}

\def\bc{{\bf c}}
\def\bz{{\bf z}}

\usepackage{ntheorem}

\newtheorem{thm}{Theorem}

\def\eg{\emph{e.g.}} 
\def\ie{\emph{i.e.}}

\usepackage{amsmath}

\def\major{\textcolor{black}}
\def\majortab{\color{black}}

\begin{document}

\title{Zero-Shot Skeleton-Based Action Recognition With Prototype-Guided Feature Alignment}
\author{Kai Zhou, Shuhai Zhang, Zeng You, Jinwu Hu, Mingkui Tan, \IEEEmembership{Senior Member, IEEE}, \\and Fei Liu, \IEEEmembership{Member, IEEE}
        % <-this % stops a space

\thanks{This work was supported by National Natural Science Foundation of China (62072190, U24A20327, U23B2013, and 62276176), and 
Guangdong Basic and Applied Basic Research Foundation (2024A1515010900). \textit{(Kai Zhou and Shuhai Zhang contributed equally to this work.) (Corresponding authors: Mingkui Tan; Fei Liu.)}}

\thanks{Kai Zhou and Fei Liu are with the School of Software Engineering, South China University of Technology, Guangzhou, China (e-mail: kayjoe0723@gmail.com, feiliu@scut.edu.cn).}

\thanks{Shuhai Zhang and Jinwu Hu are with the School of Software Engineering, South China University of Technology, and with Pazhou Lab, Guangzhou, China (e-mail: shuhaizhangshz@gmail.com, fhujinwu@gmail.com).}

\thanks{Zeng You is with the School of Future Technology, South China University of Technology, Guangzhou, China and also with Peng Cheng Laboratory, Shenzhen, China (e-mail: zengyou.yz@gmail.com).}

\thanks{Mingkui Tan is with the School of Software Engineering, South China University of Technology, Guangzhou, China, and also with the Key Laboratory of Big Data and Intelligent Robot (South China University of Technology), Ministry of Education, Guangzhou 510006, China (e-mail: mingkuitan@scut.edu.cn).}
\thanks{Code is publicly available at  \url{https://github.com/kaai520/PGFA}.}
}

\markboth{IEEE TRANSACTIONS ON IMAGE PROCESSING}% , VOL. ~, NO. ~, August~2021
{Zhou \MakeLowercase{\textit{et al.}}: Zero-Shot Skeleton-Based Action Recognition With Prototype-Guided Feature Alignment}


\maketitle

\begin{abstract}
Zero-shot skeleton-based action recognition aims to classify unseen skeleton-based human actions without prior exposure to such categories during training. This task is extremely challenging due to the difficulty in generalizing from known to unknown actions. Previous studies typically use two-stage training: pre-training skeleton encoders on seen action categories using cross-entropy loss and then aligning pre-extracted skeleton and text features, enabling knowledge transfer to unseen classes through skeleton-text alignment and language models' generalization.
However, their efficacy is hindered by 1) insufficient discrimination for skeleton features, as the fixed skeleton encoder fails to capture necessary alignment information for effective skeleton-text alignment; 2) the neglect of alignment bias between skeleton and unseen text features during testing. 
To this end, we propose a prototype-guided feature alignment paradigm for zero-shot skeleton-based action recognition, termed \textbf{\sexyname}.
Specifically, we develop an end-to-end cross-modal contrastive training framework to improve skeleton-text alignment, ensuring sufficient discrimination for skeleton features. Additionally, we introduce a prototype-guided text feature alignment strategy to mitigate the adverse impact of the distribution discrepancy during testing.
We provide a theoretical analysis to support our prototype-guided text feature alignment strategy and empirically evaluate our overall \sexyname on three well-known datasets.
Compared with the top competitor SMIE method, our \sexyname achieves absolute accuracy improvements of 22.96\%, 12.53\%, and 18.54\% on the NTU-60, NTU-120, and PKU-MMD datasets, respectively.
\end{abstract}


\section{Introduction}\label{sec:intro}
\begin{figure*}[t]
    \centering
    \includegraphics[width=1.0\textwidth]{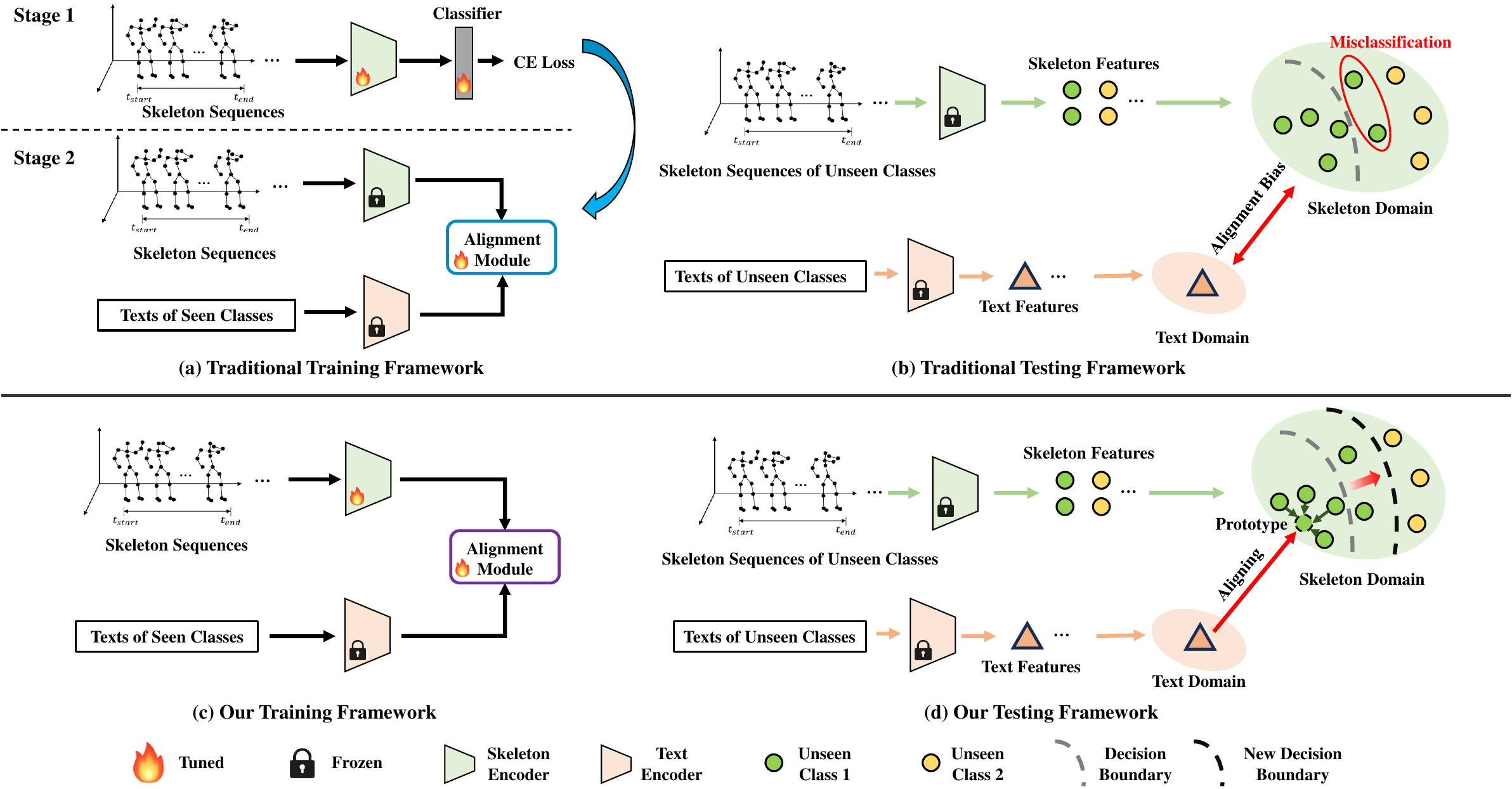}
    % \vspace{-5pt}
    \caption{\major{Illustration of traditional framework and our framework for zero-shot skeleton-based action recognition. \textbf{(a)} Traditional training framework. ``CE Loss'' denotes the cross-entropy loss. \textbf{(b)} Traditional testing framework. Unseen skeleton samples are classified by calculating similarity with text features of unseen classes (i.e., prediction based on the highest similarity score). However, due to the distributional shifts from seen to unseen classes, alignment bias between skeleton features and unseen text features is inevitable. \textbf{(c)} Our training framework. Unlike the traditional two-stage training framework, our end-to-end training framework significantly enhances training efficiency and avoids potential inconsistencies arising from independently trained components. \textbf{(d)} Our testing framework. Compared to the traditional testing framework, our testing framework leverages prototype features derived from the unseen skeleton feature distribution to replace unseen text features. It mitigates the negative impact of alignment bias between skeleton features and unseen text features.}}
    \vspace{-10pt}
    \label{fig:illustration}
\end{figure*}

\IEEEPARstart{S}{keleton-based} action recognition aims to classify human actions based on the movements of skeletal joints. It offers notable advantages in robustness to appearance variations and privacy preservation compared to RGB-based methods \cite{simonyan2014two,tran2015learning,wang2016temporal,carreira2017quo,feichtenhofer2019slowfast,tong2022videomae,li2019deep,liu2021semantics,qin2015compressive,tang2019learning,liu2023dual}, making it applicable in various domains, including human-robot interaction \cite{reily2018skeleton,bandi2021skeleton}, healthcare and rehabilitation (\eg, fall detection) \cite{yin2019skeleton,noor2023lightweight}, and sports analysis \cite{elaoud2020skeleton}.

Despite the significant progress achieved by existing fully supervised deep learning methods on this task \cite{yan2018spatial,wang2018beyond,shi2019two,zhang2019view,ye2020dynamic,hao2021hypergraph,cheng2021extremely,chen2021channel,chi2022infogcn,duan2022revisiting,zhu2022multilevel,myung2024degcn}, handling numerous action classes in real-world scenarios is uneconomical due to the high cost of annotating and labeling 3D video skeleton action data. Each annotated action clip requires costly spatiotemporal segmentation, posing practical limitations on scalability \cite{xu2017transductive}. This necessitates effective zero-shot skeleton-based action recognition methods \cite{jasani2019skeleton,gupta2021syntactically,zhou2023zero,zhu2024part} to overcome these challenges, enabling accurate action recognition without the extensive need for labeled data of new actions.

Zero-shot skeleton-based action recognition methods recognize and classify unseen human actions using skeletal data without prior exposure to such categories during training. This approach leverages knowledge transfer from previously seen actions to recognize new, unseen actions, thereby mitigating the need to collect or annotate new actions. However, this task is challenging due to the following primary difficulties. 1) The subtle differences between similar actions require highly discriminative models to distinguish them accurately \cite{zhu2024part}. 2) The substantial variance in human actions makes it difficult to generalize from known to unknown actions \cite{zhou2023zero}.  

Existing zero-shot skeleton-based action recognition methods \cite{jasani2019skeleton,gupta2021syntactically,zhou2023zero,zhu2024part} typically use a pre-trained text encoder to extract text features of action classes, which are then matched with skeleton features to predict the action class of the input skeleton sequence. This approach is adopted because text descriptions encapsulate rich semantic information that aids in generalizing to unseen action categories \cite{zhou2023zero}. Previous methods generally follow a two-stage training framework, as shown in Fig. \ref{fig:illustration}(a). Initially, they train a skeleton encoder using cross-entropy loss on data from seen action categories. Afterward, the skeleton encoder is fixed and used to extract skeleton features. Concurrently, they use a pre-trained text encoder to extract text features from class name \cite{jasani2019skeleton,zhou2023zero}, descriptions \cite{zhou2023zero,zhu2024part}, or part-of-speech tagged words \cite{gupta2021syntactically}. These text and skeleton features are used to train an alignment module to map the text and skeleton features into the same semantic feature space for seen action categories. During testing, they compute similarities between the skeleton features of test samples and the text features of unseen classes and then obtain their predictions based on the highest similarity score (see Fig. \ref{fig:illustration}(b)).
 
Despite advances in these methods, several limitations persist: 1) \textit{Insufficient discrimination in training:} The skeleton feature extractor, trained separately using the ``pretrained\&fixed" two-stage framework, fails to capture the intrinsic correlation between text and skeleton features effectively. Even with the subsequent alignment module, it struggles to produce discriminative skeleton features that align well with the textual space due to potential inconsistencies between the training modules in the two-stage process. Specifically, the skeleton encoder extracts features from unseen categories with insufficient discriminative power, resulting in a small margin between different categories (see a visualization of t-SNE in Fig. \ref{fig:feature_space}), making it difficult to match the text representations correctly. 
2) \textit{Alignment bias in testing:} During testing, substantial distributional differences between seen and unseen categories inevitably cause misalignment between skeleton features and the corresponding unseen text features, which we call alignment bias.
Ignoring these distributional differences during testing will severely hamper the model's generalization. Addressing this alignment bias to improve zero-shot prediction capability remains an important yet unresolved challenge.

To address the above issues, we investigate how to improve the alignment of the distributions of skeleton and text features from both \textit{training} and \textit{testing} perspectives.  
In the \textit{training} phase, we seek to employ an end-to-end contrastive learning approach to improve cross-modal alignment, ensuring sufficient discrimination for skeleton features. The fundamental idea behind contrastive learning is to ensure intra-class compactness by pulling together similar instances while pushing apart dissimilar ones \cite{he2020momentum}, showing great potential in cross-modal alignment tasks \cite{radford2021learning, wang2023actionclip,xue2023ulip,hegde2023clip}. 
Moreover, the end-to-end manner reduces the potential inconsistencies from independently trained components and enables efficient optimization within a single unified framework.
Benefiting from these merits, we propose \textbf{an end-to-end cross-modal contrastive training framework}. This framework avoids the limitations associated with pre-training skeleton encoders using cross-entropy loss by employing contrastive training with high-quality pre-extracted text features in an end-to-end manner. 
Additionally, we explore \major{four} action description generation methods to create diverse text inputs using pre-trained large language models. This approach aims to acquire higher-quality text features, thereby improving skeleton-text alignment and enhancing the zero-shot recognition performance.

In the \textit{testing} phase, we seek an effective prototypical learning approach to alleviate the adverse impact of the distribution discrepancy between unseen and seen action categories. Prototypical learning has proven to be a highly effective strategy for few-shot learning \cite{snell2017prototypical, huang2023location} and domain adaptation \cite{rebuffi2017icarl, lin2022prototype, iwasawa2021test}. When skeleton features of unseen classes exhibit modest intra-class compactness, selecting and creating prototype features from these skeleton features can provide better alignment with unseen skeleton features than using the original unseen text features. Since prototype features are derived from the distribution of unseen skeleton features, they better represent the central tendency of the unseen classes, leading to more consistent and robust alignment compared to isolated unseen text features. 
\major{To achieve this, we propose a \textbf{prototype-guided text feature alignment strategy}, adjusting the unseen text features to better align with the skeleton features. Initially, we obtain the pseudo-label of each skeleton sequence based on its similarity to the original unseen text features. To reduce the impact of mislabeling and noise, we introduce a filtering mechanism that selects skeleton features with high confidence, thereby improving the accuracy of subsequent prototype creation. We then create prototypes for each unseen class by calculating the centroid of the high-confidence features. Finally, we reclassify each test sample based on its similarity to the prototype feature of each unseen class. By using dynamically derived prototypes instead of static unseen text features, we address the alignment bias issue and enhance the model's generalization.}

We integrate the aforementioned training and testing frameworks into a paradigm, termed \textbf{prototype-guided feature alignment paradigm} for zero-shot skeleton-based action recognition (\textbf{\sexyname}). We qualitatively and quantitatively evaluate our \sexyname on NTU-60, NTU-120, and PKU-MMD datasets. Experimental results demonstrate that our \sexyname achieves absolute accuracy improvements of 22.96\%, 12.53\%, and 18.54\% on the NTU-60, NTU-120, and PKU-MMD datasets, respectively, compared with the top competitor SMIE \cite{zhou2023zero}. Our contributions are summarized as follows: 

\begin{itemize}
\item
We propose a prototype-guided feature alignment paradigm for zero-shot skeleton-based action recognition, termed \sexyname, to mitigate issues of insufficient discrimination and alignment bias during training and testing. Extensive experiments demonstrate the superior performance of \sexyname in zero-shot action recognition.

\item
\major{We introduce an \textbf{end-to-end} cross-modal contrastive training framework to ensure sufficient discrimination for skeleton features. Unlike traditional two-stage frameworks, our end-to-end approach enhances training efficiency and ensures intra-class compactness by aligning skeleton and text features for the same category while promoting dissimilarity for different categories. Additionally, we explore four description generation methods to improve skeleton-text alignment. Qualitative and quantitative results validate the effectiveness of our framework in enhancing skeleton feature discrimination and skeleton-text alignment.}

\item
We propose a prototype-guided text feature alignment strategy to alleviate alignment bias between skeleton and unseen text features during testing. To the best of our knowledge, we are the first to improve skeleton-text alignment during testing for zero-shot skeleton-based action recognition. Theoretical and empirical results demonstrate the effectiveness of this strategy in mitigating the negative impact of alignment bias.
\end{itemize}

\section{Related Work}
\subsection{Skeleton-based Action Recognition}
With the rise of accurate depth sensors like Kinect cameras and pose estimation algorithms, skeleton-based action recognition is gaining attention. 

In earlier times, Recurrent Neural Networks (RNNs) are applied to handle skeleton sequences \cite{du2015hierarchical, song2017end,zhang2017view}. 
Drawing inspiration from the achievements of Convolutional Neural Networks (CNNs) in image-related tasks, 2D-CNN-based methods \cite{choutas2018potion,caetano2019skelemotion,zhang2019view,xu2022topology} initially represent the skeleton sequence as a pseudo image and employ CNNs to model relationships between skeletal joints.
PoseConv3D \cite{duan2022revisiting} is the first to apply 3D-CNNs to skeleton-based action recognition. It utilizes a 3D heatmap volume to represent human skeletons, offering a novel perspective in capturing spatial-temporal dynamics of human actions. Graph Convolutional Network (GCN)-based methods \cite{yan2018spatial,shi2019two,ye2020dynamic,chen2021channel,chi2022infogcn, cheng2020skeleton,zhu2022multilevel,wang2018beyond,hao2021hypergraph,myung2024degcn,cheng2021extremely} have gained significant attention for representing human joints as graph nodes and their connections through adjacency matrices.
ST-GCN \cite{yan2018spatial} introduces a spatial-temporal graph convolutional approach to capture human joint relationships across both spatial and temporal dimensions, using predefined spatial graphs reflecting the body's natural joint connections and temporal edges for consecutive frames. 
Shift-GCN \cite{cheng2020skeleton} advances skeleton-based action recognition by leveraging shift graph convolutional networks, enabling efficient spatial and temporal joint relationship modeling without the computational burden of traditional convolutions. 
Recent interest in transformers \cite{vaswani2017attention, dosovitskiy2020image} lead to exploring transformer-based methods \cite{plizzari2021spatial, shi2020decoupled,zhang2021stst} for skeleton data. STST \cite{zhang2021stst} introduces a Spatial-Temporal Specialized Transformer, effectively modeling skeleton sequences by employing distinct joint organization strategies for spatial and temporal dimensions, capturing skeletal movements efficiently. 
In this paper, we utilize ST-GCN \cite{yan2018spatial} and Shift-GCN \cite{cheng2020skeleton} as skeleton encoders to extract features following SMIE \cite{zhou2023zero}. Our method enables the replacement of the skeleton encoder with various network architectures for training.

\subsection{Zero-shot Skeleton-based Action Recognition.} 
Despite the advancements made in skeleton-based action recognition, the existing methods \cite{jasani2019skeleton,gupta2021syntactically,zhou2023zero,zhu2024part} explored in zero-shot skeleton-based action recognition remain limited. Jasani et al. \cite{jasani2019skeleton} are the first to explore zero-shot skeleton-based action recognition. They extend DeViSE \cite{frome2013devise} and RelationNet \cite{sung2018learning} to skeleton-based action recognition by extracting skeleton features with ST-GCN pre-trained on seen classes and text embeddings
with Word2Vec \cite{word2vec} or Sentence-Bert \cite{reimers2019sentence}.  
Gupta et al. \cite{gupta2021syntactically} introduce the SynSE method, which incorporates a generative multi-modal alignment module to align skeleton features with parts of speech-tagged words. They use additional PoS syntactic information to classify labels into verbs and nouns, finding that separately extracted text embeddings enhance generalization. Their proposed SynSE outperforms their reimplemented baseline methods, including ReViSE \cite{hubert2017learning}, JPoSE \cite{wray2019fine}, and CADA-VAE \cite{schonfeld2019generalized}, in zero-shot prediction capabilities. The above methods map the skeleton or text features to fixed points in the embedding space, which does not consider the global distribution of the semantic features. To address this issue, Zhou et al. \cite{zhou2023zero} propose the SMIE method, which incorporates a global alignment module to estimate the mutual information between skeleton and text features, along with a temporal constraint module to capture the inherent temporal information of actions. Recently, Zhu et al. \cite{zhu2024part} propose the PURLS framework, which introduces a sophisticated text 
prompting module and a novel skeleton partitioning module to generate aligned textual and skeleton representations across different levels.
\major{Our method differs from previous approaches \cite{jasani2019skeleton,gupta2021syntactically,zhou2023zero,zhu2024part} in two key aspects: \textbf{(1) Training:} We use an end-to-end training framework that aligns skeleton and text features via direct cross-modal contrastive learning, unlike the commonly used two-stage training frameworks, such as SMIE \cite{zhou2023zero}, which pre-trains with cross-entropy loss and fixes the skeleton encoder. \textbf{(2) Testing:} During testing, our prototype-guided text feature alignment strategy adapts the unseen text features to better align with the skeleton features, which is a significant improvement over prior methods that rely on static unseen text features.}

\subsection{Multi-modal Representation Learning}
Recently, multi-modal representation learning methods such as CLIP \cite{radford2021learning} and ALIGN \cite{jia2021scaling} have gained significant attention for demonstrating that vision-language co-training can develop robust representations for downstream tasks, including zero-shot image classification and text-image retrieval. Subsequently, multi-modal representation learning has excelled in other fields, including video understanding \cite{wang2023actionclip,wang2024cross}, 3D point cloud understanding~\cite{zhang2022pointclip,xue2023ulip,hegde2023clip}, and 3D human action generation \cite{tevet2022motionclip}. ActionCLIP \cite{wang2023actionclip} adapts CLIP's training scheme for video action recognition by incorporating additional transformer layers into a pre-trained CLIP model for temporal modeling of video data. ULIP \cite{xue2023ulip} and CG3D \cite{hegde2023clip} enhance 3D point cloud understanding by learning a unified representation across images, texts, and point clouds. MotionCLIP \cite{tevet2022motionclip} aligns human action latent space with CLIP latent space for 3D human action generation. In the field of skeleton-based action recognition, the fully supervised GAP \cite{xiang2023generative} method first utilizes generative prompts and a multi-modal training paradigm to guide action recognition. GAP employs the alignment of skeleton and text modalities as an auxiliary task to assist supervised learning, primarily utilizing multi-part contrastive learning based on parts descriptions of the human body in motion. In contrast to GAP, the work most closely related to ours, our method mainly differs in two key aspects. First, the alignment of skeleton and text features in our method is not an auxiliary task; without the support of supervised learning, the parts descriptions used by GAP may not be suitable in zero-shot scenarios (cf. Section \ref{sec:ablation}). Second, beyond aligning skeleton and text features of seen action classes, we introduce a prototype-guided text feature alignment strategy that adjusts unseen text features for better alignment with skeleton features.

\subsection{Prototypical Learning}
Prototypical learning involves learning class prototypes to represent class features for classification tasks. It has been widely applied in few-shot learning \cite{snell2017prototypical, huang2023location}, semi-supervised learning \cite{xu2022semi}, and domain adaptation \cite{rebuffi2017icarl, lin2022prototype, iwasawa2021test}, where it effectively handles limited labeled data and evolving learning tasks by leveraging prototype-based classification. Snell et al. \cite{snell2017prototypical} propose Prototypical Networks for few-shot classification, utilizing prototype representations of each class in a learned metric space to achieve excellent results. Xu et al. \cite{xu2022semi} propose a novel approach for semi-supervised semantic segmentation that addresses intra-class variation by regularizing within-class feature distribution, leveraging consistency between linear and prototype-based predictors to encourage proximity to within-class prototypes while maintaining separation from between-class prototypes. Iwasawa et al. \cite{iwasawa2021test} propose T3A, which improves model robustness to distribution shifts by adjusting classifiers during test time using pseudo-prototype representations. Lin et al. \cite{lin2022prototype} proposed ProCA, which effectively aligns domains and preserves prior knowledge through label prototype identification and prototype-based alignment and replay strategies. However, while prototypical learning in most existing methods typically focuses on known classes, our approach concerns prototypical learning for unknown classes. 

\section{Proposed Methods}\label{sec:method}

\begin{figure*}[t]
    \centering
    \includegraphics[width=0.96\textwidth]{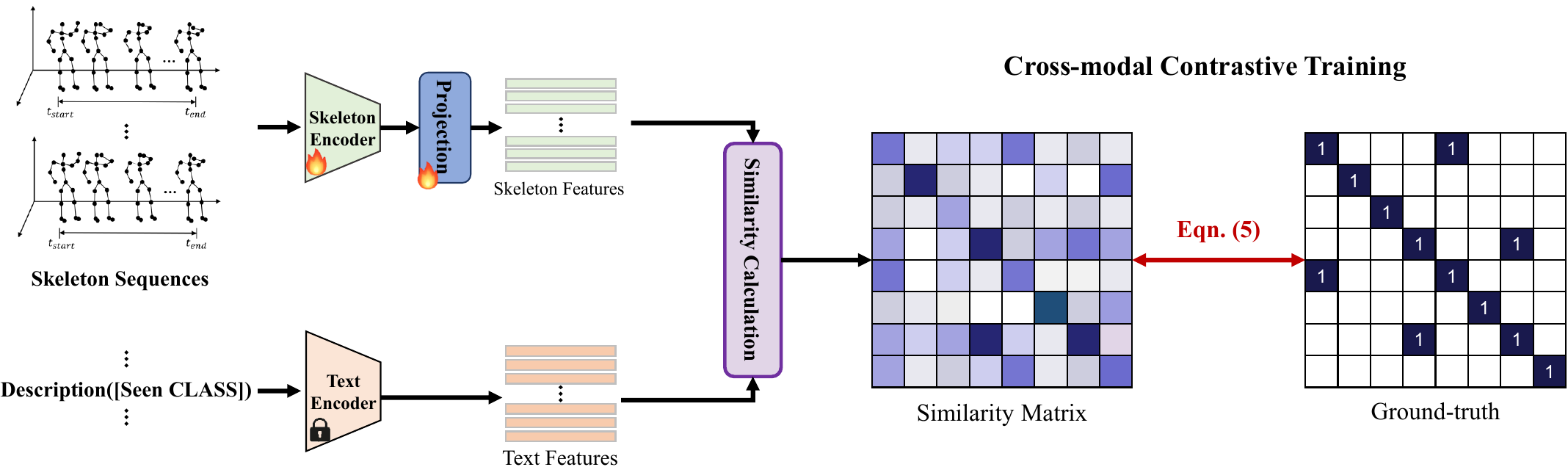}
    % \vspace{-5pt}
    \caption{Training framework of proposed \sexyname. During \textit{training}, we employ contrastive cross-modal training in an end-to-end manner to simultaneously train a skeleton encoder and a projection layer, aiming to learn a unified representation space for skeleton sequences and text.}
    % \vspace{-10pt}
    \label{fig:framework}
\end{figure*}

\subsection{Problem Formulation and Method Overview}
We aim to address zero-shot skeleton-based action recognition, where the model is trained on seen classes and tested on disjoint unseen classes. In this context, we define the training set as $\bmD_s=\{(\bx_i^s, t_i^s)\}_{i=1}^{N_s}$, where $N_s$ denotes the number of training samples, $\bx_i^s$ and $t_i^s$ denote a skeleton sequence and the corresponding action description from \textit{seen} classes, respectively. Similarly, we define the testing set as $\bmD_u=\{(\bx_i^u, t_i^u)\}_{i=1}^{N_u}$, which includes $N_u$ testing samples, with $\bx_i^u$ and $t_i^u$ representing a skeleton sequence and the corresponding action description from \textit{unseen} classes, respectively. In the zero-shot setting \cite{zhou2023zero}, \major{we have each $t_i^s \in T^s$, $t_i^u \in T^u$, and $T^s \cap T^u = \varnothing$}, where $T^s$ and $T^u$ denote action descriptions set of \textit{seen} and \textit{unseen} classes, respectively. This task is challenging due to the potential substantial variance between seen and unseen human actions, but the subtle differences between similar actions.

\begin{figure*}[t]
    \centering
    \includegraphics[width=0.96\textwidth]{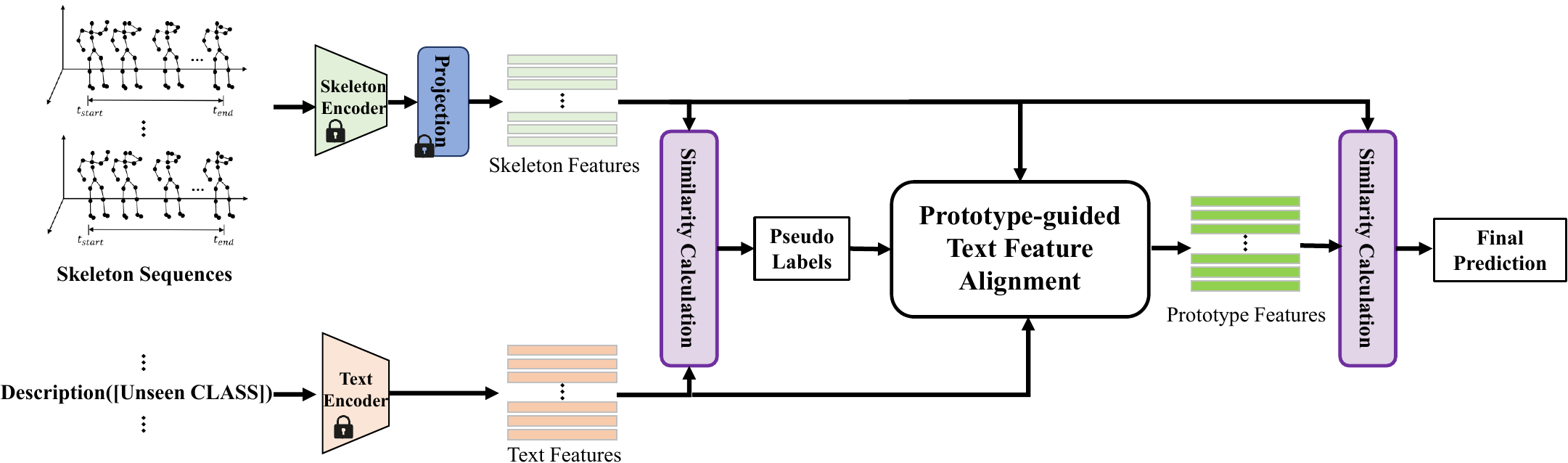}
    % \vspace{-5pt}
    \caption{Testing framework of proposed \sexyname. During \textit{testing}, we first generate pseudo-labels for each skeleton sequence based on its similarity to the original unseen text features. Using these pseudo-labels, we create a prototype feature for each unseen class. Finally, we reclassify each sample based on its similarity to the prototype feature rather than the text feature of each unseen class. For a detailed illustration of the prototype-guided text feature alignment, please refer to Fig. \ref{fig:prototype}.}
    \vspace{-10pt}
    \label{fig:testing-framework}
\end{figure*}

To address this, we aim to learn a mapping between textual and visual feature spaces to recognize new action categories and propose a prototype-guided feature alignment paradigm for zero-shot skeleton-based action recognition, called \sexyname. \sexyname consists of 1) \textbf{an end-to-end cross-modal contrastive training framework}, which learns a unified representation space of skeleton sequence and text, as illustrated in Fig. \ref{fig:framework} and Alg. \ref{alg:pipeline}, and 2) \textbf{a prototype-guided text feature alignment strategy}  for testing, which adjusts the unseen text features to align skeleton features better, as shown in Fig. \ref{fig:testing-framework} and Alg. \ref{alg:testing}.

\subsection{End-to-End Cross-modal Contrastive Training}

To construct a mapping between textual and skeletal feature spaces, existing methods \cite{jasani2019skeleton,gupta2021syntactically,zhou2023zero,zhu2024part} typically involve pre-training skeleton encoders using cross-entropy loss on seen classes, subsequently fixing them to extract features from seen skeletons, which are then matched with text features derived from a pre-trained text encoder. 
However, this two-stage training often fails to generate discriminative skeleton features that align effectively with the textual space, thereby compromising generalization performance. Additionally, numerous studies have criticized cross-entropy loss for its lack of intra-class compactness and inadequate margins \cite{liu2016large,elsayed2018large,khosla2020supervised,shi2021constrained}, which hinder subsequent modality alignment.  To address this, we introduce \textbf{an end-to-end cross-modal contrastive training framework} for zero-shot skeleton-based action recognition.

\begin{algorithm}[t]
    \renewcommand{\algorithmicrequire}{\textbf{Input:}}
    \renewcommand{\algorithmicensure}{\textbf{Output:}}
    \caption{Training \major{Scheme} for \sexyname}\label{alg:pipeline}
    \begin{algorithmic}[1]
    \REQUIRE The training dataset $\bmD_s=\{(\bx_i^s, t_i^s)\}_{i=1}^{N_s}$, the pre-trained text encoder $E_t$, the batch size $B$, the number of training epochs $T$.\\
    \ENSURE The skeleton encoder $E_x$ and the projection layer $\psi$. \\
    \FOR{$j=1,...,T$}
    \FOR{each batch $\{(\bx_i^s, t_i^s)\}_{i=1}^{B} \subset \bmD_s$}
    \STATE Calculating the training loss $\mL_{KL}$ via Eq. (\ref{loss:kl}).
    \STATE Updating the skeleton encoder $E_x$ and the projection layer $\psi$ by minimizing $\mL_{KL}$.
    \ENDFOR
    \ENDFOR
    \end{algorithmic}
\end{algorithm}

Formally, given an input skeleton sequence $\bx^s_i$ and a text description of corresponding action $t^s_i$, we first employ a skeleton encoder $E_x$ and a text encoder $E_t$ to extract the skeleton feature $\bv_i^s$  and text feature $\bw_i^s$ by:
\begin{align}
    \bv_i^{s} &= \psi(E_x(\bx^s_i)),\\
    \bw_i^{s} &= E_t(t^s_i),
\end{align}
where $\psi$ is a projection layer, ensuring the output dimension of the skeleton feature and text feature are consistent. Inspired by \cite{radford2021learning, wang2023actionclip}, we calculate the bidirectional softmax-normalized similarity scores represented by:
\begin{align}
    p_b^{x2t}(\bv_i^{s}) &= \frac{\exp(\mathrm{sim}(\bv_i^{s},\bw_b^{s})/\tau)}{\sum_{j=1}^{B}\exp(\mathrm{sim}(\bv_i^{s},\bw_j^{s})/\tau)},\\
    p_b^{t2x}(\bw_i^{s}) &= \frac{\exp(\mathrm{sim}(\bv_b^{s},\bw_i^{s})/\tau)}{\sum_{j=1}^{B}\exp(\mathrm{sim}(\bv_j^{s},\bw_i^{s})/\tau)},
\end{align}
where $\mathrm{sim}(\cdot, \cdot)$ is the cosine similarity, $\tau$ is a learnable temperature parameter and $B$ is batch size, $b$ ranges from 1 to $B$. Unlike the one-to-one image-text pairs in CLIP \cite{radford2021learning}, our task involves multiple positive matches within a batch, as several samples may belong to the same action category. It is not proper
to regard this similarity learning as a 1-in-N classification problem with InfoNCE loss \cite{oord2018representation}. Instead, we utilize Kullback–Leibler (KL) divergence as the contrastive loss for skeleton-text alignment learning:
% \mmE_{({\bf f}^x,{\bf f}^t)\sim \mB}
\begin{align}\label{loss:kl}
    \mL_{KL}{=}\frac{1}{2}\sum_{i=1}^{B}\left[\mathrm{KL}\left(\bp^{x2t}(\bv_i^{s}), {\bf m}_i^{x2t}\right){+}\mathrm{KL}\left(\bp^{t2x}(\bw_i^{s}), {\bf m}_i^{t2x}\right)\right],
\end{align}
where $\bp^{x2t}(\bv_i^{s}) {=} [p_1^{x2t}(\bv_i^{s}), \ldots, p_B^{x2t}(\bv_i^{s})]$ is the $i$-th row vector of the whole similarity scores matrix. Similarly, $\bp^{t2x}(\bw_i^{s})$ also denotes the $i$-th column vector of its similarity scores matrix. ${\bf m}_i^{x2t}$ and ${\bf m}_i^{t2x}$ represent the corresponding ground-truth similarity scores vector, which assigns a probability of 1 for positive pairs and a probability of 0 for negative pairs.

Note that $\mL_{KL}$ is exclusively employed for training the skeleton encoder $E_x$ and the projection layer $\psi$. This is necessitated by the limited number of skeleton-text pairs available for training, which is insufficient to support the training of the text encoder, unlike the ample training data used for language or vision-language pre-training \cite{reimers2019sentence,radford2021learning}. Therefore, we adopt a pre-trained large-scale language model (such as Sentence-BERT \cite{reimers2019sentence}) as our text encoder $E_t$, and we keep the parameters of the text encoder fixed during training. To save computational resources, we pre-extract text features to eliminate the need for text encoder inference during training.

\begin{figure}[tb]
    \centering
    \includegraphics[width=0.48\textwidth]{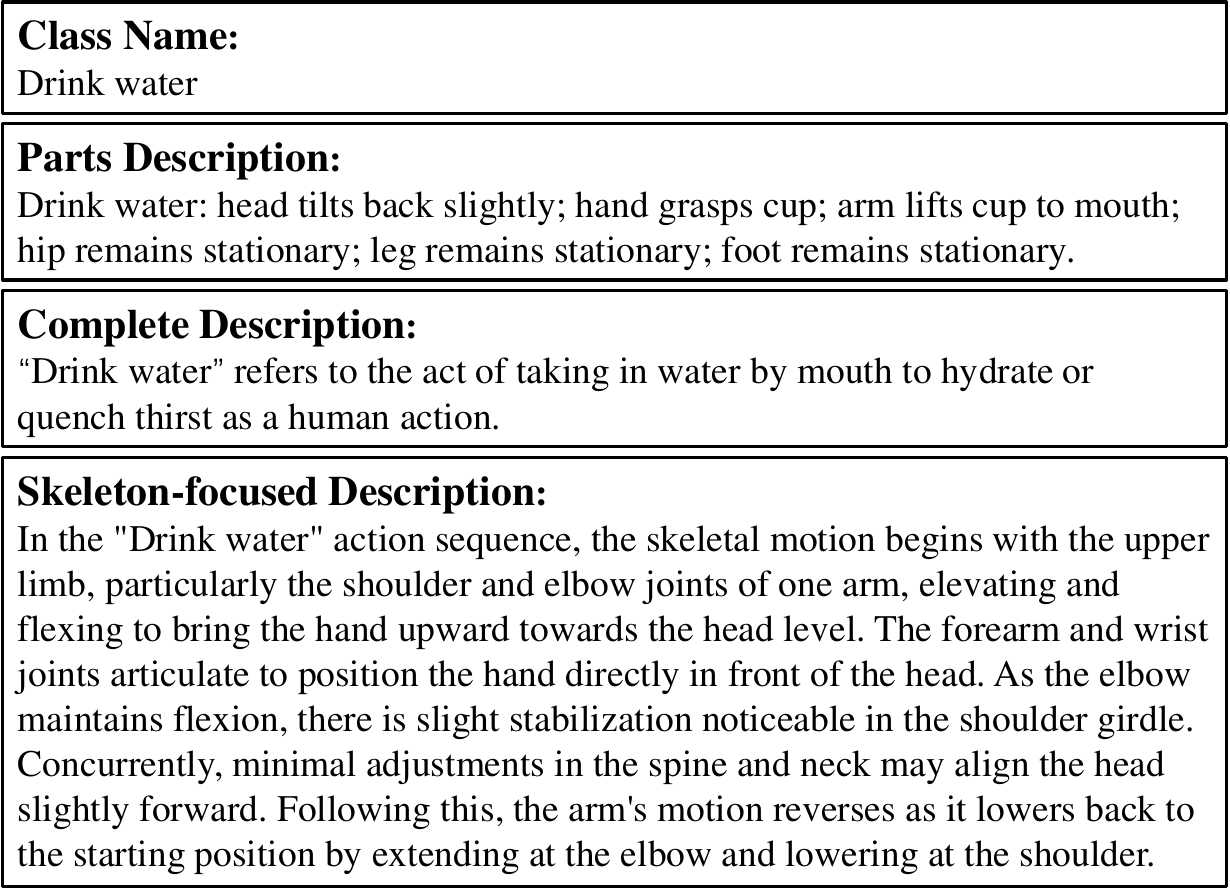}
    \vspace{-4pt}
    \caption{\major{Different descriptions of action ``drink water''.}}
    \label{fig:prompt}
    \vspace{-10pt}
\end{figure}

\textbf{Action Description Generation.} 
The quality of action description is crucial for skeleton-text alignment learning. We explore \major{four} action description generation methods in our zero-shot settings. 
In Fig. \ref{fig:prompt}, we present various generation methods for generating the action ``drink water". The details of the \major{four} methods are as follows. 

\textbf{(a) Class Name:} 
One simple and quick solution is to directly use the class name of the action as input for the text encoder. Using this generation method within our training framework serves as our baseline model. However, these action names contain only a few words and cannot fully and accurately describe the corresponding action semantics. Thus, this method leads to limited zero-shot prediction capability (as shown in our ablation studies). \textbf{(b) Parts Description:} The fully-supervised model GAP \cite{xiang2023generative} first uses generative prompts for skeleton-based action recognition. The primary action description in GAP is based on the parts description.  GAP leverages GPT-3 \cite{brown2020language} to generate action descriptions from different perspectives of body parts, including the head, hand, arm, hip, leg, and foot. Although this generation method demonstrates superior performance in supervised scenarios, its effectiveness in zero-shot scenarios is uncertain. \textbf{(c) Complete Description:} We follow the approach of SIME \cite{zhou2023zero}, expanding each class name of action into a complete action description using ChatGPT. This generation method can offer comprehensive descriptions of action semantics, which is crucial in skeleton-text alignment learning.
Finally, it is worth emphasizing that action description generation is not the primary focus of our paper. We leave ample room for further improvements, such as exploring more complex prompt engineering or prompt tuning techniques. \major{\textbf{(d) Skeleton-focused Description:} We use ChatGPT with a carefully designed prompt to ensure the output aligns closely with observable skeletal motion patterns. The prompt is as follows:
``Describe the skeletal motion pattern of a human action sequence labeled [CLASS] in a video in a single paragraph of fewer than 100 words. Focus only on joint movements, avoiding references to objects, facial expressions, or any unobservable body parts (e.g., mouth, hair, and toes). Emphasize key joint displacements, relative limb movements, and the temporal progression of the action based purely on skeletal data." This prompt guides the LLM to generate concise, motion-focused descriptions that exclude irrelevant details and emphasize joint dynamics.}

\subsection{Prototype-guided Text Feature Alignment}
In this section, we first review the standard zero-shot testing procedure and then introduce our prototype-guided text feature alignment strategy.

\textbf{Standard zero-shot testing}. 
For each test skeleton sequence $\bx^u_i$, the predicted action class $\hat{y_i}$ is determined by taking the argmax over the similarity scores with the unseen semantic features:
\begin{align}
% \label{eq:pseudo-label}
    \hat{y_i} &= \mathop{\arg\max}\limits_{k}P(\hat{y_i}=k),\label{eq:pseudo-label}\\
    P(\hat{y_i}=k) &=\frac{\exp(\mathrm{sim}(\bv^u_i,\bw^{u,k}))}{\sum_{j=1}^{K}\exp(\mathrm{sim}(\bv^u_i,\bw^{u,j}))},\label{eq:P_y}
\end{align}
where $\bv_i^u=\psi(E_x(\bx^u_i))$ denotes the skeleton feature of $\bx^u_i$, and $\bw^{u,k}=E_t(t^{u,k})$ denotes the text features of $t^{u,k}$, with $t^{u,k}\in T^u$ representing the action descriptions of the unseen class $k$. Here, $T^u=\left\{t^{u,1}, \ldots, t^{u,K}\right\}$ is a set of action descriptions for unseen classes, consisting of $K$ action descriptions. In the zero-shot setting, $\bw^{u,k}$ is not involved in training, so there is no guarantee that it will align well with unseen skeleton features.

\textbf{Prototype-guided text feature alignment strategy.}
To address the above issue, benefitting from the effectiveness of prototypical learning \cite{rebuffi2017icarl, lin2022prototype,iwasawa2021test} in transfer learning scenarios, we propose \textbf{a prototype-guided text feature alignment strategy} to better align with unseen skeleton features in a zero-shot setting.
Specifically, we first acquire the pseudo label $\hat{y_i}$ of each skeleton sequence $\bx_i^u$ based on Eq. (\ref{eq:pseudo-label}). Then we create a prototype feature for each unseen class to replace the corresponding unseen text feature, using skeleton features whose pseudo-labels belong to that unseen class. To this end, we create a support set 
$\mmS^k$ for each unseen class $k$:
\begin{align}\label{eq:Sk}
    \mmS^k=\left\{ \frac{\bv^u_i}{||\bv^u_i|| }\mid \hat{y_i}=k,i \leq N_u,i\in\mmN^+\right\},
\end{align}
where $||\cdot||$ denotes the L2 norm. $N_u$ denotes the number of testing samples in the testing set. Intuitively, taking the centroid of the support set $\mmS^k$ as a prototype feature to replace the unseen text feature $\bw^{u,k}$ can align well with the skeleton features belonging to the unseen class k. 
However, inevitably, some pseudo-labels are misassigned to incorrect classes, making their utilization undesirable due to the introduction of noise into the prototype feature. To tackle this issue, we use the prediction entropy $H(\bz) = -\sum_k P(\hat{y_i}=k\mid\bz)\log P(\hat{y_i}=k\mid\bz)$ to filter the element $\bz = \frac{\bv^u_i}{||\bv^u_i||}$ in $\mmS^k$. To be specific, we create a final support set $\mmZ^k$ for each seen class $k$:
\begin{align}\label{eq:Zk}
    \mmZ^k = \left\{ \bz\mid\bz \in \mmS^k,H(\bz)\leq h^k\right\},
\end{align}
where $h^k$ is the $\rho^k$-th smallest entropy of the support set $\mmS^k$, with $\rho^k = \lfloor \alpha \cdot |\mmS^k| \rfloor$ denoting the size of $\mmZ^k$ and $\alpha \in [0,1]$ being a tolerance margin. We can modify the size of $\mmZ^k$ by the hyperparameter $\alpha$. The impact of $\alpha$ is illustrated in Fig. \ref{fig:alpha}. We define a prototype feature to replace $\bw^{u,k}$ as follows:
\begin{align}\label{eq:prototype}
\bc^{u,k}=\left\{
\begin{aligned}
    \frac{1}{|\mmZ^k|}\sum_{\bz\in\mmZ^k}\bz&,&if\ \mmZ^k \neq \varnothing, \\
    \bw^{u,k}&,&otherwise,
\end{aligned}
\right.
\end{align}
where $\bc^{u,k}$ is the prototype feature of unseen class $k$. Note that even if there exist no features belonging to $\mmZ^k$, \ie, $\mmZ^k = \varnothing$, the centroid of $\mmZ^k$ can be reduced to the original text feature $\bw^{u,k}$.
Built upon the prototype features, We reclassify each test skeleton sequence $\bx^u_i$ by:
\begin{align}\label{eq:final}
    \tilde{y_i} =\mathop{\arg\max}\limits_{k}\frac{\exp(\mathrm{sim}(\bv^u_i,\bc^{u,k}))}{\sum_{j=1}^{K}\exp(\mathrm{sim}(\bv^u_i,\bc^{u,j}))},
\end{align}
where $\tilde{y_i}$ denotes the final predicted action class of $\bx^u_i$.

\begin{algorithm}[t]
    \renewcommand{\algorithmicrequire}{\textbf{Input:}}
    \renewcommand{\algorithmicensure}{\textbf{Output:}}
    \caption{Testing \major{Scheme} for \sexyname}\label{alg:testing}
    \begin{algorithmic}[1]
    \REQUIRE The testing dataset $\bmD_u$, the text encoder $E_t$, the skeleton encoder $E_x$ and the projection layer $\psi$.\\
    \ENSURE The final prediction set $\{\tilde{y_i}\}_{i=1}^{N_u}$. \\
    \FOR{each $\bx_i^u \in \bmD_u$}
    \STATE Generating a pseudo label $\hat{y_i}$ of $\bx_i^u$ via Eq. (\ref{eq:pseudo-label}).
    \ENDFOR
    \FOR{each unseen class $k=1,...,K$}
    \STATE Calculating a prototype feature $\bc^{u,k}$ via Eq. (\ref{eq:prototype}).
    \ENDFOR
    \FOR{each $\bx_i^u \in \bmD_u$}
    \STATE Generating a final predicted label $\tilde{y_i}$ of $\bx_i^u$ via Eq. (\ref{eq:final}).
    \ENDFOR
    \end{algorithmic}
\end{algorithm}

\textbf{Theoretical analysis for prototype-guided text feature alignment strategy.}
To elucidate the mechanism of our proposed alignment strategy, we would like to analyze it from a statistical view, making it easier to understand the core insights of this method. To achieve this, we provide the following theorem to give an intuitive explanation, where we do not consider the filter operation on the support set $\mmZ^k$ and $\mmZ^k \neq \varnothing$. For simplicity, we denote the normalized skeleton feature of the $k$-th class as $\bv^{(k)}$, where $\|\bv^{(k)}\|=1$. The normalized features are constrained to lie on the unit hypersphere. Therefore, we assume that the features follow a von Mises–Fisher distribution \cite{banerjee2005clustering, wood1994simulation}, which is considered a generalization of the normal distribution to the unit hypersphere. This distribution's advantageous property in representing cosine similarity by dot-products simplifies theoretical analysis.

\begin{thm}
\label{thm: argP}
    Assuming that the distributions of the $K$ classes of normalized skeleton feature $\bv^{(k)}$ are from $K$ von Mises–Fisher distributions \cite{banerjee2005clustering} with the same concentration parameter $\kappa$ but different mean directions $\boldsymbol{\mu}_k$, i.e., $f_{p_k}(\bv;\boldsymbol{\mu}_k,\kappa)=C_{d}(\kappa)\exp(\kappa\boldsymbol{\mu}_k^\mathsf{T}\mathbf{v})$, $k=1, \ldots, K$, where $C_{d}(\kappa)$ is a normalization constant related to the concentration parameter $\kappa$ and feature dimension $d$, 
    and the $k$-th class of prototype feature is $\hat{\boldsymbol{\mu}}_k= \frac{\frac{1}{n}\sum_{i=1}^n \bv_i^{(k)}}{\|\frac{1}{n}\sum_{i=1}^n \bv_i^{(k)}\|} $, which is a normalized $\bc^{u,k}$ in Eq. (\ref{eq:prototype}).
    Then, as $n \to \infty$, for $\forall~\bv \sim p_m$, $m=1,\ldots K$, if $k=\mathop{\arg\max}\limits_{i}\frac{\exp(\mathrm{sim}(\bv,\hat{\boldsymbol{\mu}}_i))}{\sum_{j=1}^{K}\exp(\mathrm{sim}(\bv,\hat{\boldsymbol{\mu}}_j))}$, we have
    \begin{equation}
        P(\bv | \mathrm{class}~k)>P(\bv | \mathrm{class}~j), \forall~j \neq k.
    \end{equation}
\end{thm}

\begin{figure}[tb]
    \centering
    \includegraphics[width=0.49\textwidth]{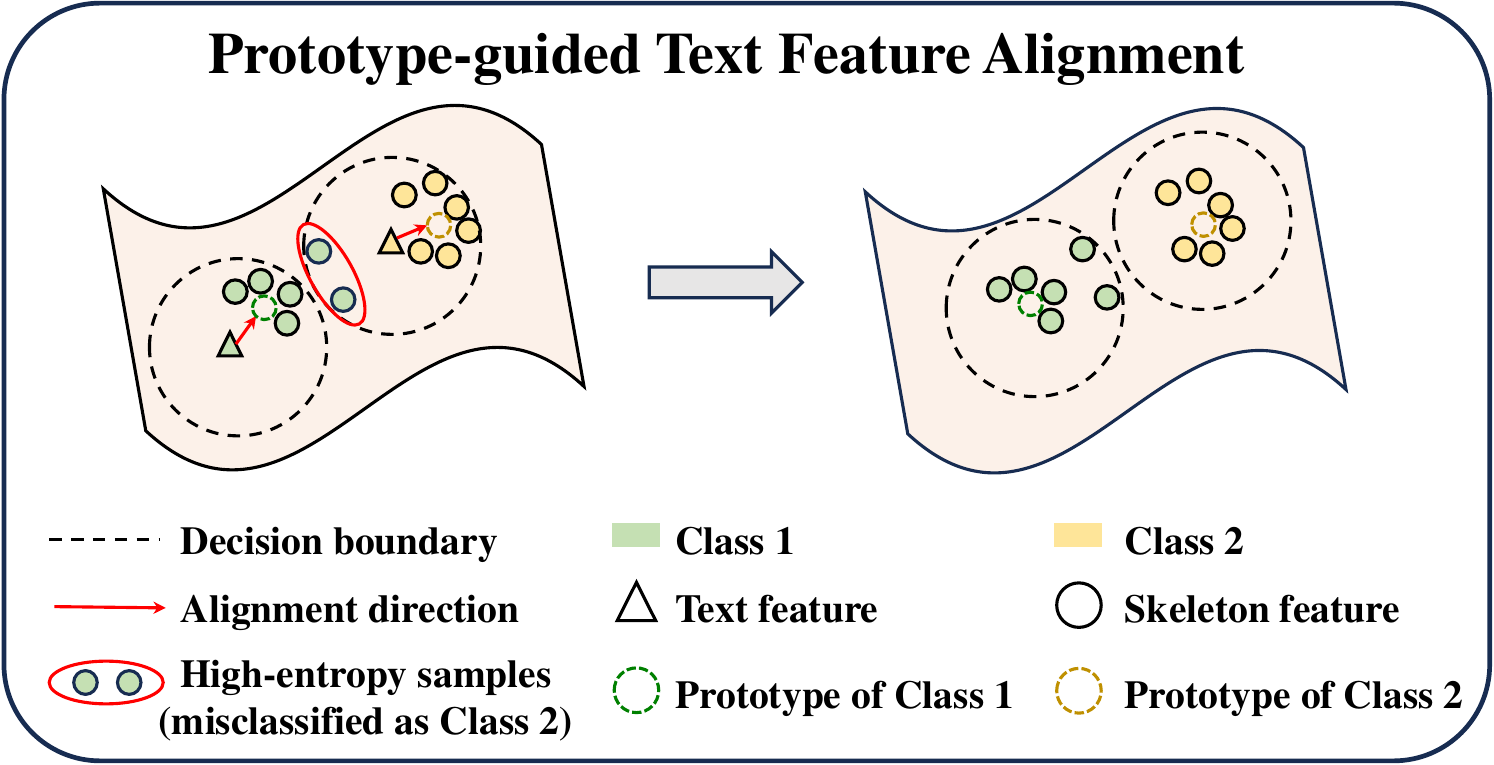}
    % \vspace{-10pt}
    \caption{Illustration of prototype-guided text feature alignment strategy. Class 1 and Class 2 represent two unseen classes. We illustrate the alignment directions for text features within the feature space before alignment. After alignment, we use the prototype features to replace the original text features in constructing the decision boundaries.}
    % \vspace{-10pt}
    \label{fig:prototype}
\end{figure}

\textit{For the proof of Theorem \ref{thm: argP}, please refer to the supplementary.} Theorem \ref{thm: argP} shows that when the number of samples $n$ in $\mmZ$ is large, if a sample has the highest similarity to the class center vector of a particular class, then the probability that it belongs to this class is higher than that of it belonging to other classes. Therefore, reclassification according to the maximum similarity criterion in Eq. (\ref{eq:final}) can improve action classification accuracy from a statistical view. Note that a large $n$ results in a more accurate class center, whereas a small $n$ makes the class center more sensitive to noise. Our entropy-based filtering mechanism mitigates this issue, as demonstrated by the experiments in Fig. \ref{fig:alpha}.

\section{Experiments}
\subsection{Datasets}
\textbf{NTU-RGB+D 60} \cite{shahroudy2016ntu} comprises 56,578 skeleton sequences belonging to 60 action categories, which were performed by 40 subjects. These skeleton sequences were captured using Microsoft Kinect sensors, with each subject being represented by 25 joints. In supervised settings, the dataset is usually split into training and testing sets based on subjects or views. However, in zero-shot settings, this dataset is divided into training and test sets based on action categories.

\textbf{NTU-RGB+D 120} \cite{liu2019ntu} is an extended dataset of NTU-60 (short for NTU-RGB+D 60), encompassing data from 106 subjects and comprising 113,945 skeleton sequences across 120 action categories. In our zero-shot settings, this dataset is also divided into training and test sets based on action categories.

\textbf{PKU-MMD} \cite{liu2020benchmark}  dataset consists of action samples across 51 categories, collected from 66 subjects. The data is captured using Kinect v2 sensors from multiple viewpoints. This dataset is divided into two parts: Part I includes 21,539 samples, and Part II includes 6,904 samples. 
We utilize Part I to split the training and test sets according to action categories in this setting, following the SMIE method \cite{zhou2023zero}.

\subsection{Evaluation Settings}

\textbf{Setting \uppercase\expandafter{\romannumeral1}.}
Due to the substantial influence of various class splits on zero-shot results, SMIE \cite{zhou2023zero} suggests a setting on 3 datasets for further verifying the stability of different methods. 
To achieve this, a three-fold test is applied to each dataset to reduce variance. Each fold comprises distinct groups of seen and unseen classes, and the \textbf{average accuracy} of the predicted unseen classes is reported.
For NTU-60, they offer three groups of 55/5 class splits, with each group comprising 55 seen and 5 unseen classes. For NTU-120, they provide three groups of 110/10 class splits, with each group comprising 110 seen and 10 unseen classes. For PKU-MMD, they offer three groups of 46/5 class splits, with each group comprising 46 seen and 5 unseen classes.
For a fair comparison, we employ ST-GCN \cite{yan2018spatial} as our skeleton encoder and Sentence-BERT \cite{reimers2019sentence} as our text encoder in this setting, following the SMIE method.

\textbf{Setting \uppercase\expandafter{\romannumeral2}.}
Additionally, we also offer a more extensive comparison with another evaluation setting proposed by SynSE \cite{gupta2021syntactically}. This setting maintains two fixed class partitions on NTU-60 and NTU-120 datasets. Specifically, for the NTU-60 dataset, SynSE presents 55/5 and 48/12 splits, encompassing 5 and 12 unseen classes, respectively. For the NTU-120 dataset, SynSE offers 110/10 and 96/24 splits. In contrast to Setting \uppercase\expandafter{\romannumeral1}, a one-fold test is required for each class split, rather than a three-fold test. The \textbf{accuracy} is reported. We employ Shift-GCN \cite{cheng2020skeleton} as our skeleton encoder and Sentence-BERT \cite{reimers2019sentence} as our text encoder, following the approach used in SynSE.

\subsection{Implementation Details}
In Setting \uppercase\expandafter{\romannumeral1}, we adopt the identical data processing procedure as Cross-CLR \cite{li20213d}, following SMIE. This procedure entails eliminating invalid frames and adjusting the skeleton sequences to a length of 50 frames through linear interpolation. The skeleton feature extractor employed is ST-GCN with 16 hidden channels, resulting in an extracted feature dimension of 256. We utilize Sentence-BERT to acquire the 768-dimensional text feature following SMIE guidelines. Hence, the projection layer $\psi$ is implemented as a linear layer, projecting from 256 to 768 dimensions. For all experimental runs, we employ the SGD optimizer with 50 epochs. The learning rate is set to $5\times 10^{-2}$ for the NTU-60 and PKU-MMD datasets, whereas for the NTU-120 dataset with a larger data size, the learning rate is adjusted to $5\times 10^{-3}$ to ensure smoother training. We opt for the complete description of action classes as the default action description due to its superior performance. The tolerance margin $\alpha$, which controls the size of final support sets, is set to 0.9, 0.4, and 1.0 for the NTU-60, NTU-120, and PKU-MMD datasets, respectively. The effect of $\alpha$ will be analyzed in ablation studies.

In Setting \uppercase\expandafter{\romannumeral2}, apart from utilizing Shift-GCN as the skeleton encoder, all other implementation details remain consistent with Setting \uppercase\expandafter{\romannumeral1}.  Notably, SynSE employs 4s-Shift-GCN pre-trained on seen classes to extract skeleton features. The 4s-Shift-GCN means utilizing the joint stream, bone stream, joint-motion stream, and bone-motion stream of skeleton sequences to train 4 Shift-GCN models with cross-entropy loss, and finally averages the outputs of these 4 Shift-GCN models. However, our method does not require a pre-trained skeleton encoder with cross-entropy loss, while employing four encoders in our end-to-end training paradigm would be overly complex. Therefore, we only use the joint stream with one Shift-GCN model to train our method, which may lead to relatively weaker skeleton feature extraction but improve training efficiency.

\subsection{Main Results}
\begin{table}[tb]

  \begin{center}
  \caption{Comparsion with previous methods on NTU-60, NTU-120, and PKU-MMD datasets under Setting \uppercase\expandafter{\romannumeral1}. SMIE$^\dag$ indicates that it utilizes the complete description of action classes as text input. \sexynameplus$^*$ refers to our method utilizing the class name as text input but not employing our prototype-guided text feature alignment strategy. \sexynameplus$^\dag$ refers to our method utilizing the complete description but not employing our prototype-guided text feature alignment strategy. \sexyname denotes our full method.}
  \label{tab:main}
  % \vspace{-5pt}
  % \resizebox{\linewidth}{!}{
  \begin{tabular}{c|c|c|c}
    \toprule
    Method & NTU-60(\%) & NTU-120(\%) & PKU-MMD(\%)\\
    Split: (Seen/Unseen)        & (55/5) & (110/10) & (46/5)\\
    \midrule
     DeViSE \cite{frome2013devise} & 49.80 & 44.59 & 47.94 \\
     RelationNet \cite{sung2018learning} & 48.16 & 40.55 & 51.97 \\
     ReViSE \cite{hubert2017learning} & 56.97 & 49.32 & 65.65 \\
     SMIE \cite{zhou2023zero} &63.57 & 56.37 & 67.15 \\
     SMIE$^\dag$\cite{zhou2023zero} & 70.21 & 58.85 & 69.26 \\
     \midrule
     \sexynameplus$^*$(Ours) & 65.17 & 57.56& 74.76\\
     \sexynameplus$^\dag$(Ours) & 80.45 & 60.22& 78.34\\
     \sexyname(Ours)  & \textbf{93.17} & \textbf{71.38}& \textbf{87.80}\\
    \bottomrule
  \end{tabular}%}
  \end{center}
  % \vspace{-10pt}
\end{table}

Under Setting \uppercase\expandafter{\romannumeral1}, we compare our \sexyname with previous methods, following the comparison methods outlined in SMIE \cite{zhou2023zero}. Previous methods commonly utilize the class name of action classes as input to a text encoder. SMIE additionally provides experimental results using the complete description of actions as input.  For fair comparisons, we provide \sexynameplus$^*$ and \sexynameplus$^\dag$, which use class names and complete descriptions as input, respectively, but do not incorporate our prototype-guided text feature alignment strategy. Additionally, we offer a version incorporating our prototype-guided text feature alignment strategy as our full method to validate its performance. All the datasets get 3 groups of class splits provided by SMIE and the average accuracies are reported on Tab. \ref{tab:main}. As shown in Tab. \ref{tab:main}, both \sexynameplus$^*$ and \sexynameplus$^\dag$ demonstrate enhancements compared to the top competitor SMIE, especially with \sexynameplus$^\dag$ notably surpassing SMIE$^\dag$ by a large margin. 
This indicates that higher-quality action descriptions better unlock the potential of our end-to-end training framework for achieving improved skeleton-text alignment compared to the previous two-stage training framework. By integrating our prototype-guided text feature alignment strategy, our full method \sexyname achieves average accuracies surpassing SMIE$^\dag$ on the NTU-60, NTU-120, and PKU-MMD datasets by 22.96\%, 12.53\%, and 18.54\%, respectively. The larger performance gains by our full method demonstrate the effectiveness of our prototype-guided text feature alignment strategy.

Under Setting \uppercase\expandafter{\romannumeral2}, we compare our \sexyname with previous methods, following the comparison methods outlined in SynSE \cite{gupta2021syntactically}, SMIE \cite{zhou2023zero}, and PURLS \cite{zhu2024part}. 
Apart from using a single Shift-GCN in our end-to-end training framework instead of 4s-Shift-GCN, all other experimental settings are identical to those used in SynSE. Despite the potential for relatively weaker skeleton feature extraction in our \sexyname, \sexyname still achieves state-of-the-art performance. In the 55/5 and 48/12 splits of the NTU-60 dataset, our \sexyname surpasses the top competitor PURLS \cite{zhu2024part} by 1.03\% and 15.00\% (relatively 1.30\% and 36.59\%), respectively. Compared to our method, PURLS shows similar performance in the 55/5 split, which can be attributed to using various complex description prompts and a more powerful text encoder of CLIP \cite{radford2021learning} instead of Sentence-BERT \cite{reimers2019sentence}. In the 110/10 and 96/24 splits of the NTU-120 dataset, our \sexyname surpasses PURLS by 8.04\% and 7.41\% (relatively 11.17\% and 14.25\%), respectively. As the number of unseen classes increases, aligning skeleton and text modalities becomes increasingly challenging. Other methods performed the worst in the 48/12 split of the NTU-60 dataset. In contrast, our method achieved the largest relative improvement in this split (e.g., relatively 36.59\% compared with PURLS), demonstrating promising potential in skeleton-text alignment.

\begin{table}[tb]
\begin{center}
  \caption{Comparsion with previous methods on NTU-60 and NTU-120 datasets under Setting \uppercase\expandafter{\romannumeral2}.}
  % \vspace{-5pt}
\begin{tabular}{c|cc|cc}
\toprule
Method        & \multicolumn{2}{c|}{NTU-60(\%)} & \multicolumn{2}{c}{NTU-120(\%)} \\
Split: (Seen/Unseen) & (55/5)      & (48/12)      & (110/10)      & (96/24)     \\
\midrule
DeViSE \cite{frome2013devise}       & 60.72       & 24.51        & 47.49         & 25.74       \\
RelationNet \cite{sung2018learning}   & 40.12       & 30.06        & 52.59         & 29.06       \\
ReViSE \cite{hubert2017learning}      & 53.91       & 17.49        & 55.04         & 32.38       \\
JPoSE \cite{wray2019fine}        & 64.82       & 28.75        & 51.93         & 32.44       \\
CADA-VAE \cite{schonfeld2019generalized}     & 76.84       & 28.96        & 59.53         & 35.77       \\
SynSE \cite{gupta2021syntactically}        & 75.81       & 33.30        & 62.69         & 38.70       \\
SMIE \cite{zhou2023zero}          & 77.98       & 40.18        & 65.74         & 45.30       \\
PURLS \cite{zhu2024part} & 79.23      & 40.99       & 71.95         & 52.01       \\
\midrule
\sexyname(Ours)      & \textbf{80.26}       & \textbf{55.99}        & \textbf{79.99}         & \textbf{59.42}      \\
\bottomrule
\end{tabular}
\end{center}
% \vspace{-10pt}
\label{tab:mainv2}
\end{table}

\subsection{Ablation Studies}
\label{sec:ablation}
We conducted experiments to evaluate the effect of the key components and hyper-parameters of our method. To ensure the stability of key components and hyper-parameters, we conduct ablation studies under Setting \uppercase\expandafter{\romannumeral1} suggested by SMIE.

\begin{table}[tb]
\tabcolsep=12pt
  \begin{center}
  \caption{\major{Effect of action description generation. We explore the effect of action descriptions on skeleton-text alignment training, without using prototype-guided text feature alignment of our method.}}
   \label{tab:description}
  % \vspace{-5pt}
  % \resizebox{\linewidth}{!}{
  \begin{tabular}{c|c|c}
    \toprule
    Action Description & NTU-60(\%) & NTU-120(\%) \\
    \midrule
    Class Name & 65.17 & 57.56\\
    Parts Description & 75.65 & 52.73\\
    Complete Description & \textbf{80.45} & 60.22\\
    \major{Skeleton-focused Description}& \major{77.25} & \major{\textbf{62.22}} \\
    \bottomrule
  \end{tabular}%}
  \end{center}
  % \vspace{-10pt}
\end{table}

\begin{table}[t]
% \belowrulesep=0pt
% \aboverulesep=0pt
\tabcolsep=3pt
  \begin{center}
  \caption{\major{Effect of action descriptions generated by different LLMs on zero-shot skeleton-based action recognition. All models do not use the prototype-guided text feature alignment strategy.}}
   \label{tab:llm}
   {\majortab
  \begin{tabular}{c|c|c|c}
    \toprule
    Action Description & LLM &NTU-60(\%) & NTU-120(\%) \\
    \midrule
    \multirow{3}{*}{Complete Description}  & MiniCPM3 & 71.60& 52.29\\
    & DeepSeek-V3 & 78.75 & 59.08\\
    & ChatGPT & \textbf{80.45} & \textbf{60.22}\\
    \midrule
    \multirow{3}{*}{Skeleton-focused Description} & MiniCPM3 & 76.59& 52.42\\
    & DeepSeek-V3 & 75.41& 57.26\\
    & ChatGPT & \textbf{77.25} & \textbf{62.22}\\
    \bottomrule
  \end{tabular}}
  \end{center}
  % \vspace{-10pt}
\end{table}

% \noindent
\major{\textbf{Effect of action description generation.}} In skeleton-text alignment, without pre-training skeleton encoders with cross-entropy, the quality of pre-extracted text features becomes crucial for enhancing zero-shot capability.
\major{We explore the effect of four action description generation methods and validate these effects on the NTU-60 and NTU-120 datasets. Since GAP \cite{xiang2023generative} does not provide parts descriptions for PKU-MMD, we are concerned that our self-generated parts descriptions may differ in style. Therefore, we do not perform validation on the PKU-MMD dataset.}
As shown in Tab. \ref{tab:description}, utilizing the class name of the action as input indeed results in limited performance, primarily because it fails to describe the corresponding action semantics accurately. While parts description performs well in supervised scenarios \cite{xiang2023generative}, it is not as effective as the complete description in zero-shot scenarios. The possible reason is that parts descriptions for some actions are too similar. For instance, actions like ``put on a shoe" and ``take off a shoe" differ only in the description of the foot part, such as ``foot inserts into shoe" and ``foot grasps the shoe and pulls it off", while descriptions of other body parts remain largely the same. In supervised scenarios, parts description serves as auxiliary supervision. However, in zero-shot scenarios, the text features generated by actions like ``put on a shoe" and ``take off a shoe" exhibit high similarity, which may introduce interference in primary skeleton-text alignment. In contrast, using complete descriptions directly is more effective than parts descriptions in such cases. 
\major{The results in Tab.~\ref{tab:description} show that both Complete Description and Skeleton-focused Description outperform the other two methods. On the NTU-60 dataset, using complete descriptions performs better, while on the NTU-120 dataset, using skeleton-focused descriptions yields superior results. This may be due to the larger size and more classes in NTU-120, which demand better alignment with the skeleton data, needing Skeleton-focused Description to more accurately reflect key joint movements for the more complex actions. Considering the fairness of comparison with other zero-shot methods, we primarily use Class Name and Complete Description for comparison.}
Additionally, employing more complex prompt engineering or prompt tuning techniques might further improve results. However, this aspect is not the primary focus of our paper and will be left for future exploration. 

\begin{table*}[tb]
\tabcolsep=10pt
  
  \begin{center}
  \caption{Effect of our end-to-end training framework and prototype-guided text feature alignment. Note that ``End-to-End" is our default training framework. ``Feature Alignment" denotes our prototype-guided text feature alignment strategy.}
  \label{tab:training_paradigm}
  % \vspace{-5pt}
  % \resizebox{\linewidth}{!}{
  \begin{tabular}{cccc|c|c|c}
    \toprule
    \multicolumn{3}{c}{Training Framework} & Testing Strategy & NTU-60(\%) & NTU-120(\%) & PKU-MMD(\%)\\
    Pretraining\&Fixed & Pretraining\&Finetuning & End-to-End & Feature Alignment & (55/5) & (110/10) & (46/5)\\
    \midrule
     \checkmark& & & &73.65& 56.12& 71.85\\
     & \checkmark& & &76.99& 57.55& 74.63\\
     & & \checkmark& &80.45& 60.22& 78.34\\
     \midrule
     \checkmark& & & \checkmark & 87.23 & 68.66& 80.39 \\
     & \checkmark& & \checkmark  & 88.55 & 68.74& 82.48\\
     & & \checkmark & \checkmark  &\textbf{93.17}& \textbf{71.38}& \textbf{87.80}\\
    \bottomrule
  \end{tabular}%}
  \end{center}
  % \vspace{-10pt}
\end{table*}

\major{\textbf{Effect of action descriptions generated by different LLMs.} We explore the impact of action descriptions generated by different LLMs in our zero-shot settings. We select the open-source models MiniCPM3 \cite{hu2024minicpm} and DeepSeek-V3 \cite{liu2024deepseek}, as well as the closed-source model ChatGPT, to generate descriptions. As shown in Tab.~\ref{tab:llm}, descriptions generated by ChatGPT consistently achieve the best performance on both NTU-60 and NTU-120 datasets, for both complete and skeleton-focused descriptions. In contrast, MiniCPM3, with its lightweight 4B parameters, demonstrates a performance gap relative to the other two models. This emphasizes the crucial role of LLM quality in improving zero-shot recognition accuracy.}

\textbf{Effect of our end-to-end training framework.} To further investigate the effects of different training frameworks, we assess the zero-shot prediction capability of three frameworks on our method. \textbf{1) Pretraining\&Fixed.} This represents the training framework of previous methods, which involves initially pre-training a skeleton encoder with a cross-entropy loss using data from seen categories. Subsequently, the skeleton encoder is fixed, and alignment with the text modality is performed. \textbf{2) Pretraining\&Finetuning.} This framework involves initially pre-training a skeleton encoder with a cross-entropy loss but finetuning the skeleton encoder for subsequent skeleton-text alignment. \textbf{3) End-to-End.} This represents our end-to-end training framework, which involves directly training the skeleton encoder for skeleton-text alignment. To ensure fairness and leverage the superiority of complete descriptions, all models use complete descriptions as text input. As shown in Tab. \ref{tab:training_paradigm}, the Pretraining\&Fixed framework performs the worst for our method, indicating that the fixed skeleton feature distribution from pretraining the skeleton encoder with cross-entropy loss is not conducive to subsequent skeleton-text alignment. By allowing finetuning of the skeleton encoder during subsequent skeleton-text alignment, the Pretraining\&Finetuning framework naturally performs better than the Pretraining\&Fixed framework. For our method, the End-to-End framework proves to be the most effective across all three datasets, highlighting its superiority. Pretraining the skeleton encoder is unnecessary, as it yields suboptimal performance and incurs additional pretraining costs. 

\textbf{Effect of our prototype-guided text feature alignment.} To further evaluate the effectiveness and stability of our prototype-guided text feature alignment strategy, we assess its performance across three different training frameworks of our method on three datasets. As shown in Tab. \ref{tab:training_paradigm}, employing our prototype-guided text feature alignment strategy in the testing phase yields remarkable improvements across all training frameworks and datasets, achieving an absolute accuracy improvement of approximately 10\%. This demonstrates the strategy's effectiveness and stability.

\begin{figure}[tb]
    \centering
    \includegraphics[width=0.47\textwidth]{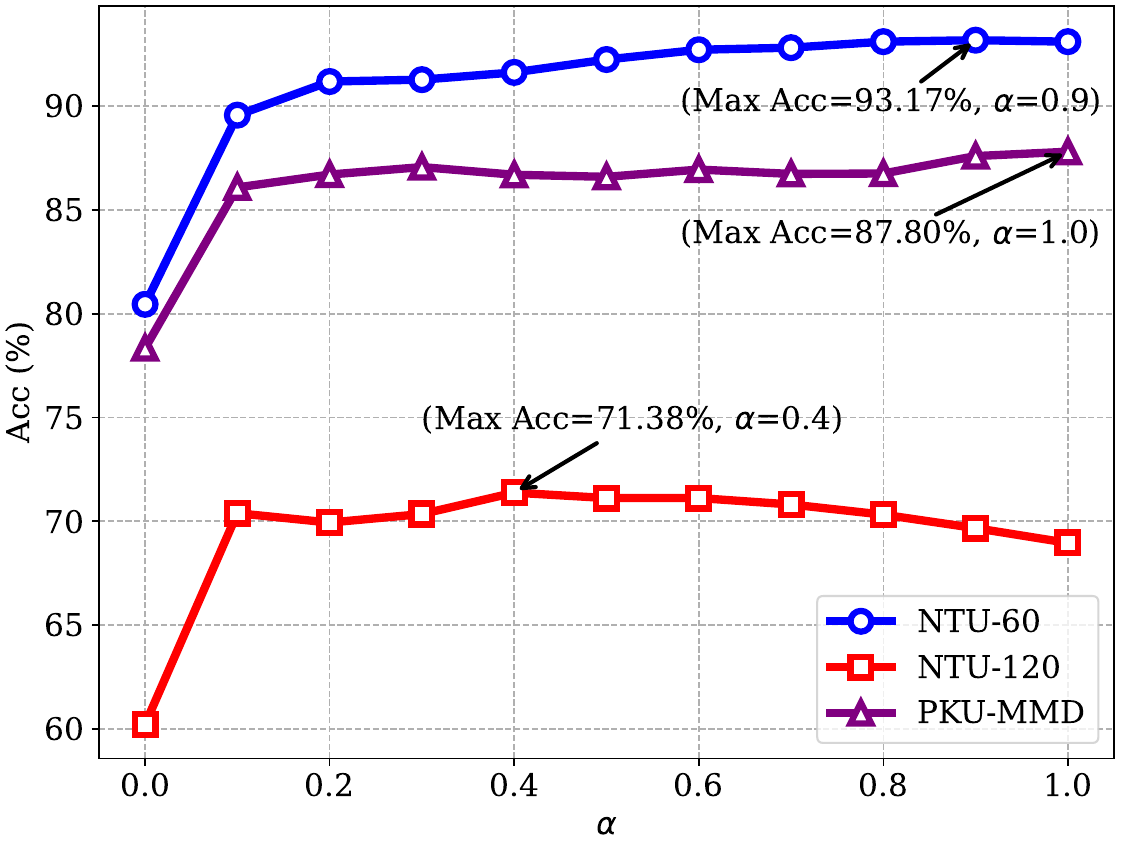}
    % \vspace{-10pt}
    \caption{Effect of tolerance margin $\alpha$. The x-axis represents the value of $\alpha$, while the y-axis represents the average accuracy for the three class splits under Setting \uppercase\expandafter{\romannumeral1}.}
    % \vspace{-10pt}
    \label{fig:alpha}
\end{figure}

\textbf{Effect of the tolerance margin $\alpha$.} In our prototype-guided text feature alignment strategy, the tolerance margin $\alpha$ controls the size of the final support set $\mmZ^k$. All models use complete descriptions as text input for training. Fig. \ref{fig:alpha} shows the performance for different values of $\alpha$. 
\major{When $\alpha=0$, the prototype-guided text feature alignment strategy is not utilized, resulting in significantly lower baseline performance. When $\alpha=1$, it means that we use all elements of support set $\mmS^k$ to construct $\mmZ^k$. Critically, for $\alpha \in [0.1, 1.0]$, our method achieves stable performance ($>95\%$ of peak accuracy) across all datasets while consistently surpassing the $\alpha=0$ baseline by large margins (e.g., NTU-120 attains 71.38\% at $\alpha=0.4$ vs. 60.22\% baseline). Intuitively, setting $\alpha$ too low may restrict the prototype calculation to few low-entropy elements, introducing bias, while overly high $\alpha$ risks including misclassified samples.}
Certainly, the optimal $\alpha$ may vary across different datasets. For NTU-120, the best performance is achieved with $\alpha=0.4$. Perhaps due to the high accuracy achieved by our method on the NTU-60 and PKU-MMD datasets, it performs optimally when $\alpha=0.9$ and $\alpha=1.0$, requiring only a small portion of elements with high prediction entropy to be filtered out.
\major{However, empirical results reveal robustness: deviations from optimals (e.g., NTU-120’s $\alpha=0.4$ vs. $\alpha=0.9$ yielding 98\% of peak accuracy) maintain near-optimal performance, highlighting adaptability to data distributions. Optimal $\alpha$ varies slightly (e.g., 0.9 for NTU-60 vs. 0.4 for NTU-120), yet $\alpha \geq 0.1$ ensures significant and stable improvements without requiring fine-tuning, underscoring the strategy’s reliability for zero-shot generalization.}

\begin{table}[t]
% \belowrulesep=0pt
% \aboverulesep=0pt
\tabcolsep=4.5pt
  \begin{center}
  \caption{\major{Effect of different pseudo-labeling strategies on prototype generation.}}
   \label{tab:pseudo}
  % \vspace{-5pt}
  % \resizebox{\linewidth}{!}{
  {\majortab
  \begin{tabular}{c|c|c|c}
    \toprule
    Pseudo-labeling Strategy & NTU-60(\%) & NTU-120(\%) & PKU-MMD(\%)\\
    \midrule
    weighted & 82.90 & 57.75& 81.51\\
    argmax (default) &\textbf{93.17}& \textbf{71.38}& \textbf{87.80}\\ 
    \bottomrule
  \end{tabular}}%}
  \end{center}
  % \vspace{-10pt}
\end{table}

\major{\textbf{Effect of different pseudo-labeling strategies on prototype generation.} In our prototype-guided text feature alignment strategy, the pseudo-labels used for prototype generation are typically derived through argmax classification. We explored an alternative approach where the pseudo-labels are weighted by their probability. Specifically, we propose to modify the calculation of the prototype feature of class $k$ as follows: $\mathbf{c}^{u,k}=\frac{\sum^{|\bmD_u|}_{i=1}P^{k}_i\cdot(\mathbf{v}^u_i/||\mathbf{v}^u_i||)}{\sum^{|\bmD_u|}_{i=1}P^{k}_i},$ where $|\bmD_u|$ is the total number of samples in the test set, $P^{k}_i$ represents the probability of the pseudo-label for sample $i$ belonging to class $k$, and $\mathbf{v}^u_i$ is the corresponding skeleton feature. As shown in Tab. \ref{tab:pseudo}, the performance of the weighted pseudo-labeling strategy is generally lower than the default argmax-based approach. This could be due to the inclusion of features from many samples that do not belong to the correct class, introducing noise into the prototype calculation. Specifically, when probability weighting is used, low-probability samples still contribute to the prototype, which is especially problematic when the model’s classification performance is weak. For example, the NTU-120 dataset shows a significant performance drop with the weighted strategy (57.75\%) compared to the argmax approach (71.38\%).}

\begin{table}[t]
\tabcolsep=20pt
% \belowrulesep=0pt
% \aboverulesep=0pt
  \begin{center}
  \caption{\major{Computational cost comparison between our method and baseline SMIE. \sexynameplus$^\dag$ refers to our method without the prototype-guided text feature alignment strategy.}}
  \label{tab:comp}
  % \vspace{-5pt}
  % \resizebox{\linewidth}{!}{
  {\majortab
  \begin{tabular}{c|c|c}
    \toprule 
    Method  & Param &  FLOPs (G)\\
    \midrule\
    SMIE \cite{zhou2023zero} & 2.38M & 0.58\\
    \sexynameplus$^\dag$(Ours) & 1.00M & 0.57\\
    \bottomrule
  \end{tabular}}%}
  \end{center}
  % \vspace{-10pt}
\end{table}

\major{\textbf{Computational complexity analysis.} We compare the computational cost of our method and the baseline SMIE in terms of parameters and floating point operations per second (FLOPs). As shown in Tab.~\ref{tab:comp}, our method significantly reduces the number of parameters to 1.00M, compared to SMIE’s 2.38M. This is because, with the same ST-GCN backbone, our method adds only one extra linear layer, while SMIE adds three fully connected layers. Without the prototype-guided text feature alignment strategy, our method’s FLOPs are also slightly lower than SMIE's. With the feature alignment strategy, FLOPs double as each sample is processed twice during inference.}

\subsection{\major{Extension Experiments}}

\begin{table}[t]
% \tabcolsep=10pt
% \belowrulesep=0pt
% \aboverulesep=0pt
  \begin{center}
  \caption{\major{Evaluation on one-shot skeleton-based action recognition. The table shows accuracy (\%) on unseen classes.}}
  \label{tab:one-shot}
  % \vspace{-5pt}
  % \resizebox{\linewidth}{!}{
  {\majortab
  \begin{tabular}{c|c|c|c}
    \toprule 
    Method  & NTU-60(\%) & NTU-120(\%) & PKU-MMD(\%)\\ 
    Split: (Seen/Unseen)        & (50/10) & (100/20) & (41/10)\\
    \midrule\
    ProtoNet \cite{snell2017prototypical} & 74.8 & 60.4 & 78.1 \\
    FEAT \cite{ye2020few} & 74.3 & 61.5 & 75.9 \\
    Subspace \cite{simon2020adaptive} & 75.6 & 60.9 & 75.6 \\
    Dynamic Filter \cite{xu2021learning} & 75.9 & 60.6 & 78.8 \\
    M\&C-scale \cite{yang2024one} & 82.7 & 68.7 & \textbf{86.9} \\
     \sexyname(Ours)  & \textbf{83.8} & \textbf{69.8}& \textbf{86.9}\\
    \bottomrule
  \end{tabular}}
  \end{center}
  % \vspace{-10pt}
\end{table}

\major{\textbf{Evaluation on one-shot skeleton-based action recognition.} To further evaluate the generalizability of our method, we extend it to the one-shot skeleton-based action recognition setting, where each unseen class is associated with only one labeled skeleton sample during testing. In our framework, adapting to the one-shot scenario is straightforward. Specifically, during testing, we use the available one-shot skeleton samples to generate class prototypes using our skeleton encoder and feature alignment strategy, instead of relying solely on text descriptions. Following \cite{yang2024one}, we evaluate on the NTU-60, NTU-120, and PKU-MMD datasets, with seen/unseen class splits of 50/10, 100/20, and 41/10, respectively. \textit{For detailed dataset splits, please refer to the supplementary.} As shown in Tab. \ref{tab:one-shot}, our \sexyname demonstrates competitive or superior performance compared to existing one-shot approaches \cite{yang2024one,snell2017prototypical,ye2020few,simon2020adaptive,xu2021learning}, highlighting its flexibility and effectiveness in one-shot scenarios.}

\begin{table*}[t]
\tabcolsep=15pt
% \belowrulesep=0pt
% \aboverulesep=0pt
  \begin{center}
  \caption{\major{Exploring self-supervised models in zero-shot scenarios. The evaluation is based on our Setting I. \sexynameplus$^\dag$ refers to our method utilizing the complete description but not employing our prototype-guided text feature alignment strategy.}}
  \label{tab:self}
  % \vspace{-5pt}
  % \resizebox{\linewidth}{!}{
  {\majortab
  \begin{tabular}{c|c|c|c|c}
    \toprule 
    Method  & \multirow{2}{*}{Training Framework} & NTU-60(\%) & NTU-120(\%) & PKU-MMD(\%)\\ 
    Split: (Seen/Unseen)  &      & (55/5) & (110/10) & (46/5)\\
    \midrule\
    Supervised \cite{yan2018spatial} & Pretraining\&Fixed &73.65& 56.12& 71.85\\
    Supervised \cite{yan2018spatial} & Pretraining\&Finetuning &76.99& 57.55& 74.63 \\
    Skeleton-logoCLR \cite{hu2024global} & Pretraining\&Fixed &75.01& 55.59& 73.27\\
    Skeleton-logoCLR \cite{hu2024global} & Pretraining\&Finetuning &79.30& 56.57& 77.33\\
    \sexynameplus$^\dag$(Ours) & End-to-End & \textbf{80.45} & \textbf{60.22} & \textbf{78.34}\\
    \bottomrule
  \end{tabular}}%}}
  \end{center}
  % \vspace{-10pt}
\end{table*}

\major{\textbf{Exploring self-supervised models in zero-shot scenarios.} We explore the potential of self-supervised models for zero-shot skeleton-based action recognition. To this end, we re-implement the state-of-the-art self-supervised model, Skeleton-logoCLR \cite{hu2024global}, replacing the supervised pre-trained skeleton encoder used in the ``Pretraining\&Fixed" and ``Pretraining\&Finetuning" training frameworks. The key distinction between these frameworks is whether the pre-trained skeleton encoder is fixed during the subsequent skeleton-text alignment. As shown in Tab.~\ref{tab:self}, under the ``Pretraining\&Fixed" framework, using Skeleton-logoCLR for pretraining outperforms using cross-entropy loss for supervised pertaining on average across three datasets, suggesting that self-supervised training may produce more discriminative features. In the ``Pretraining\&Finetuning" framework, using Skeleton-logoCLR also naturally yields better performance on average. However, our method, which uses the End-to-End framework, achieves the best performance across all datasets. This suggests that supervised and self-supervised pretraining may not be necessary in zero-shot settings. Our End-to-End training framework enables effective skeleton-text alignment while avoiding the additional cost of pretraining.}

\begin{table}[t]
% \tabcolsep=10pt
% \belowrulesep=0pt
% \aboverulesep=0pt
  \begin{center}
  \caption{\major{Cross-dataset evaluation. ``NTU-60$\to$PKU-MMD'' represents training on NTU-60 and testing on PKU-MMD, while ``PKU-MMD$\to$NTU-60'' represents training on PKU-MMD and testing on NTU-60.}}
  \label{tab:cross}
  % \vspace{-5pt}
  % \resizebox{\linewidth}{!}{
  {\majortab
  \begin{tabular}{c|c|c}
    \toprule 
    Method  & NTU-60$\to$PKU-MMD(\%)&  PKU-MMD$\to$NTU-60(\%)\\
    \midrule\
    SMIE \cite{zhou2023zero} & 75.07 & 55.92\\
    \sexyname(Ours) & \textbf{94.73} & \textbf{61.02}\\
    \bottomrule
  \end{tabular}}%}
  \end{center}
  \vspace{-10pt}
\end{table}
\major{\textbf{Cross-dataset evaluation of our method.} We further evaluate the generalization of our method through cross-dataset experiments. Since NTU-120 is an extension of NTU-60, using them together for cross-dataset evaluation is less meaningful. Instead, we perform evaluations between NTU-60 and PKU-MMD. Specifically, we train on one dataset and test directly on the other without fine-tuning: (1) training on NTU-60 and testing on PKU-MMD, and (2) training on PKU-MMD and testing on NTU-60. To further validate the stability of our method under cross-dataset settings, we follow a protocol similar to Setting I. We train on three groups of 55 seen classes from NTU-60 (Class Split 1/2/3) and test on three groups of 5 unseen classes from PKU-MMD (Class Split 1/2/3), and vice versa—training on 46 seen classes from PKU-MMD and testing on 5 unseen classes from NTU-60. We report the average accuracy on the predicted unseen classes across all splits. As shown in Tab.~\ref{tab:cross}, our method outperforms the baseline SMIE \cite{zhou2023zero} by a significant margin. Specifically, our method achieves an impressive 94.73\% average accuracy (NTU-60$\to$PKU-MMD), even surpassing performance from training on seen classes in PKU-MMD. This could be due to the larger training set in NTU-60, which enables the model to learn more discriminative features. Conversely, the smaller training set in PKU-MMD leads to slightly lower performance when transferring to NTU-60, though our method still outperforms the SMIE baseline. These results strongly demonstrate the transferability of our method.}

\subsection{Qualitative Analysis}
\begin{figure}[tb]
    \centering
    \includegraphics[width=0.47\textwidth]{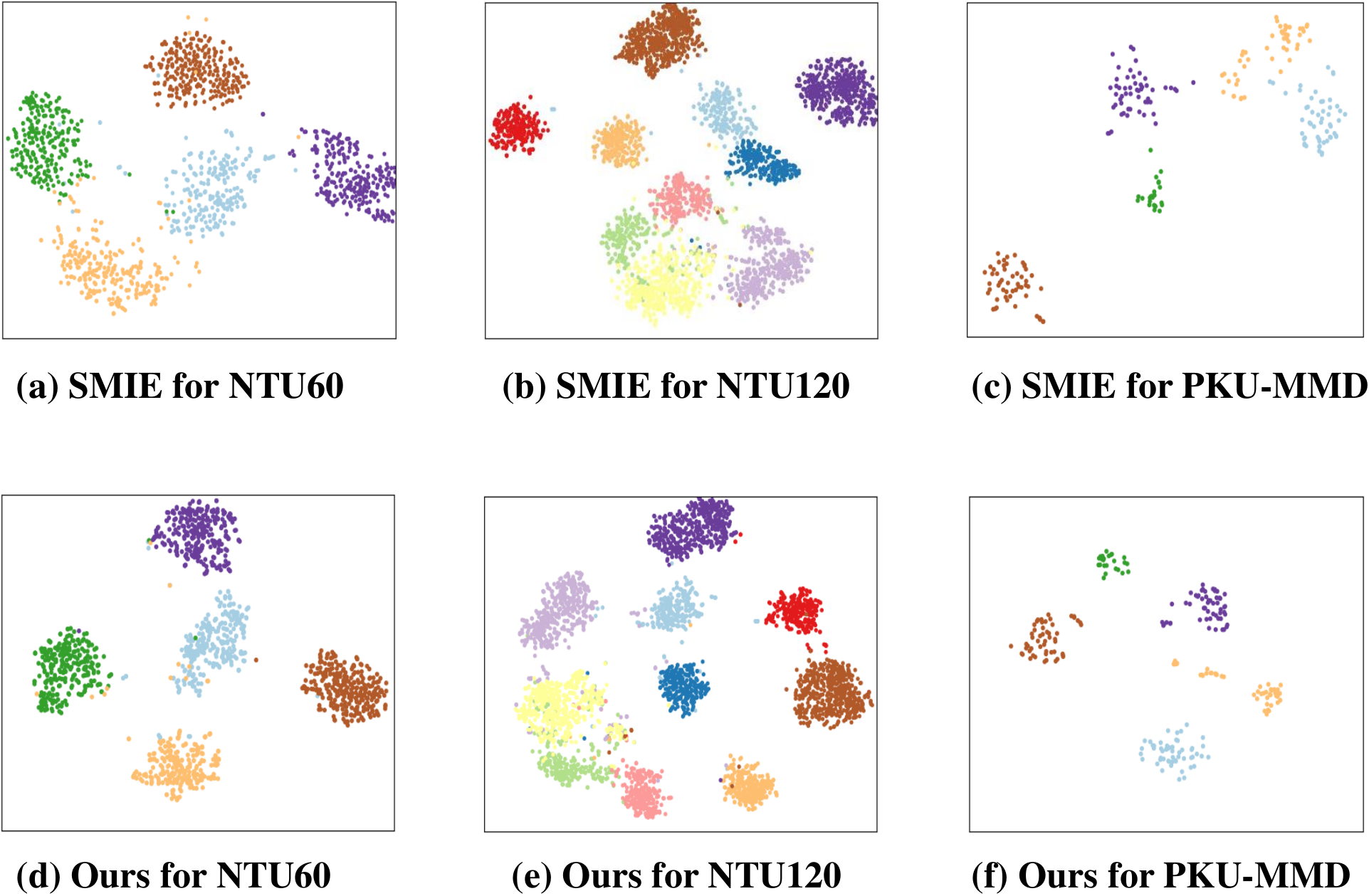}
    % \vspace{-5pt}
    \caption{
    \major{
    Visualization of skeleton features using t-SNE \cite{van2008visualizing} from the skeleton encoder pre-trained with cross-entropy loss in SMIE \cite{zhou2023zero} and the skeleton encoder of our method. These encoders are trained on data from seen classes in NTU-60, NTU-120, and PKU-MMD under Setting \uppercase\expandafter{\romannumeral1}. We visualize the skeleton features from unseen classes.}
}
    \vspace{-10pt}
    \label{fig:feature_space}
\end{figure}

\textbf{Visualization of skeleton feature space.} Existing methods always pre-train skeleton encoders using cross-entropy loss before modality alignment. However, this approach does not ensure the intra-class compactness \cite{shi2021constrained} of skeleton features, which poses challenges for subsequent skeleton-text alignment. To verify this, we compared the t-SNE \cite{van2008visualizing} visualizations of skeleton features between SMIE and our end-to-end training framework. \major{As illustrated in Fig. \ref{fig:feature_space},} the intra-class compactness of skeleton features from the existing training framework is inadequate. The use of cross-entropy loss for pre-training, which leads to a lack of intra-class compactness and the possibility of poor margins, has been analyzed in existing works\cite{liu2016large,elsayed2018large,khosla2020supervised,shi2021constrained}. The fixed distribution of such skeleton features presents challenges in skeleton-text alignment. In contrast, our training framework improves intra-class compactness through cross-modal contrastive training. Additionally, our method simplifies the entire training process without the need to pre-train the skeleton encoder.

\begin{figure}[t]
    \centering
    \includegraphics[width=0.49\textwidth]{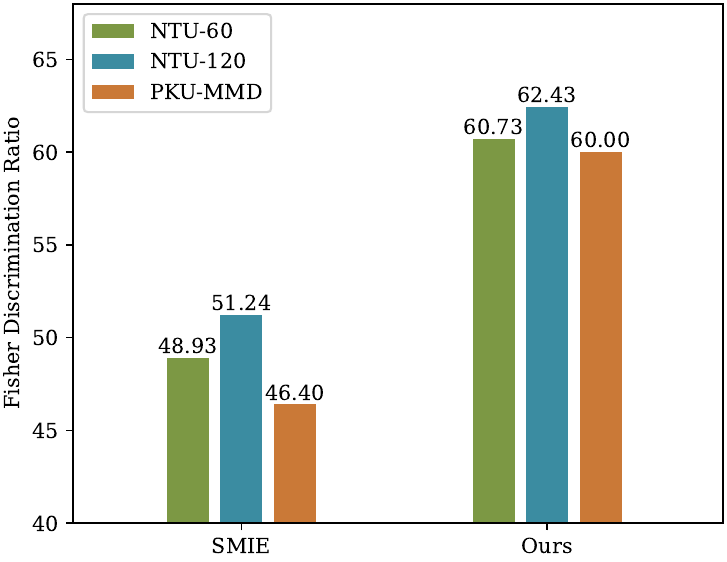}
    % \vspace{-5pt}
    \caption{\major{The average Fisher Discrimination Ratio (FDR) of skeleton features from unseen classes across 3 class splits under Setting \uppercase\expandafter{\romannumeral1}. A higher FDR indicates better class separability.}}
    \label{fig:fisher}
    \vspace{-5pt}
\end{figure}

\major{\textbf{Fisher discrimination ratio of skeleton features.} We use the Fisher Discrimination Ratio (FDR) \cite{mclachlan2005discriminant} to further assess class separability. FDR measures how well-separated classes are based on their feature distributions. For multi-class classification, FDR is calculated as the ratio of the between-class scatter to the within-class scatter: $\text{FDR}=\operatorname{tr}(S_w^{-1}S_b)$, where $S_w$ is the within-class scatter matrix, $S_b$ is the between-class scatter matrix, and $\operatorname{tr}(\cdot)$ is the trace operator. The $S_w$  is computed as the sum of the covariance matrices for each class, while $S_b$ captures the variance between the class means and the overall mean. A higher FDR indicates better class separability. For each dataset (NTU-60, NTU-120, and PKU-MMD), we calculate the average FDR of skeleton features from unseen classes across three class splits under Setting \uppercase\expandafter{\romannumeral1}. As shown in Fig. \ref{fig:fisher}, our method consistently demonstrates higher FDRs than SMIE across all three datasets, indicating better class separability and discrimination in our approach.}

\begin{figure}[t]
    \centering
    \includegraphics[width=0.49\textwidth]{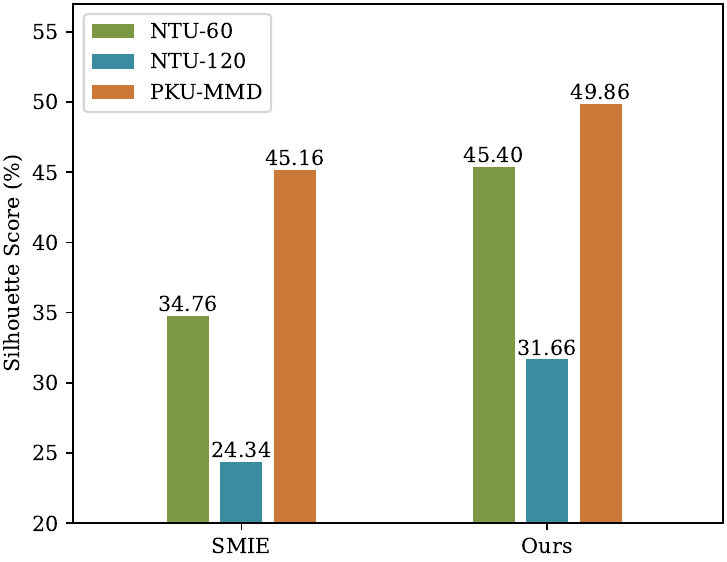}
    % \vspace{-5pt}
    \caption{\major{The average silhouette scores \cite{rousseeuw1987silhouettes} of unseen skeleton features across 3 class splits under Setting \uppercase\expandafter{\romannumeral1}. The silhouette score closer to 1 (\ie, 100\%) indicates well-clustered data, reflecting denser clusters with minimal within-cluster distance.}}
    \label{fig:ss}
    \vspace{-5pt}
\end{figure}

\major{\textbf{Silhouette score of skeleton features.} To better demonstrate the clustering differences of unseen skeleton features between SMIE (all previous methods using the same pre-trained skeleton encoder) and our method, we utilize the silhouette score \cite{rousseeuw1987silhouettes} to measure the intrinsic structure of the created clusters. The silhouette score is defined as $S=\frac{1}{N}\sum_i^N \frac{b_i-a_i}{\max(a_i,b_i)}$, where $N$ is the number of samples, $a_i$ is the average distance between sample $i$ and all other samples within the same cluster, and $b_i$ is the average distance from $i$ to all samples in the nearest neighboring cluster. The silhouette score $S$ ranges from -1 to 1. The $S$ closer to 1 indicates well-clustered data, reflecting denser clusters with minimal within-cluster distance. In this case, we use cosine distance for computation. For each dataset (NTU-60, NTU-120, and PKU-MMD), we calculate the average silhouette score for skeleton features from unseen classes across three class splits under Setting \uppercase\expandafter{\romannumeral1}. The cluster labels correspond to the true unseen class labels. As shown in Fig. \ref{fig:ss}, our method consistently demonstrates higher silhouette scores than SMIE across all datasets. It demonstrates the effectiveness of our training framework in clustering unseen skeleton features, indicating that our method better captures the inherent structure and similarities within the skeleton data.}

\begin{figure}[tb]
    \centering
    \includegraphics[width=0.49\textwidth]{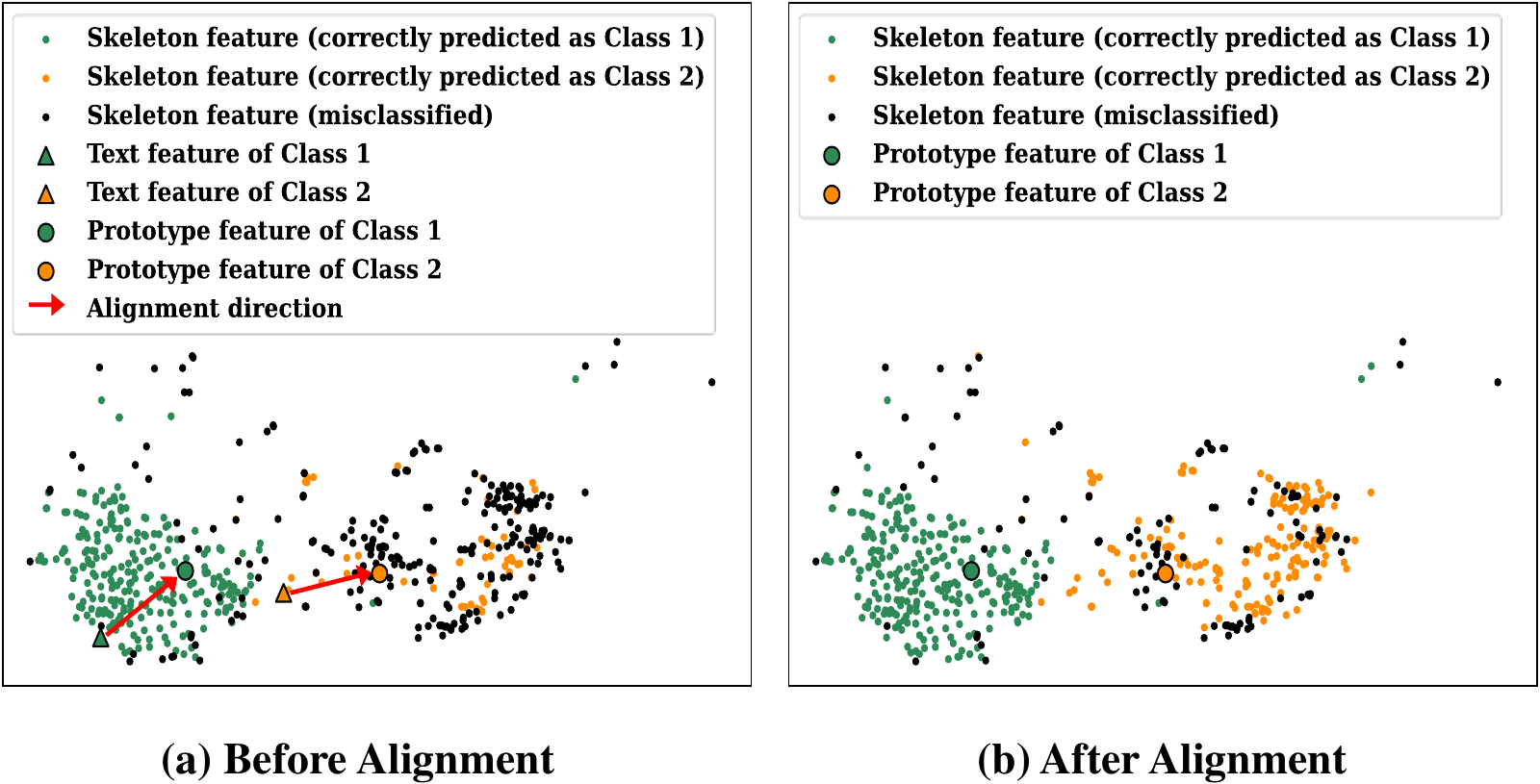}
    % \vspace{-5pt}
    \caption{Visualization of prototype-guided text feature alignment in feature space. Class 1 and Class 2 represent the unseen action categories ``Typing Keyboard" and ``Check Time", respectively. We visualize the skeleton features of these two unseen classes and their corresponding text features using t-SNE \cite{van2008visualizing}. Through our prototype-guided text feature alignment strategy, many previously misclassified samples are correctly predicted as Class 2.}
    \vspace{-10pt}
    \label{fig:alignment}
\end{figure}

\textbf{Visualization of prototype-guided text feature alignment in feature space.} To further demonstrate the effectiveness of our prototype-guided text feature alignment strategy, we visualize this strategy in the feature space. To enhance the clarity of the visualization, we select the unseen classes ``Typing Keyboard" and ``Check Time" from Class Split 2 of NTU-60 under Setting \uppercase\expandafter{\romannumeral1} \textit{(for the full confusion matrices, providing a more intuitive view of each class accuracy, please refer to the supplementary)}. The skeleton features of these two unseen classes and their corresponding text features are projected into a 2D space using t-SNE \cite{van2008visualizing}, as shown in Fig. \ref{fig:alignment}(a). Through our prototype-guided text feature alignment strategy, we obtain the prototype features for the corresponding unseen classes. These prototype features are used in place of text features for similarity calculations with skeleton features to produce the final predictions. As shown in Fig. \ref{fig:alignment}(b), many misclassified samples are corrected. This demonstrates that our prototype-guided text feature alignment strategy can effectively utilize the distribution of unseen skeleton features to derive prototype features and rectify classification errors.

\section{Discussion and Conclusion}
In this paper, we propose a prototype-guided feature alignment (\sexyname) paradigm to address issues of insufficient discrimination and alignment bias during training and testing in previous zero-shot skeleton-based action recognition methods. Our \sexyname paradigm comprises: 1) an end-to-end contrastive training framework to ensure sufficient discrimination for skeleton features, and 2) a prototype-guided text feature alignment strategy to alleviate alignment bias between skeleton and unseen text features during testing. Experimental results on NTU-60, NTU-120, and PKU-MMD datasets demonstrate the superior performance of \sexyname in zero-shot action recognition.

\textbf{Future work.} Although our prototype-guided text feature alignment strategy is highly effective, it requires classifying all test samples twice, making it unsuitable for real-time online scenarios where models cannot know all test samples in advance.  Modifications may be necessary for such scenarios. For instance, creating a continuously updated ``prototype bank" could allow for real-time updates of prototype features by calculating the predicted label of each single or batch sample during testing.
We believe this is a promising and feasible direction and plan to explore it in future work.

\nolinenumbers

\bibliographystyle{IEEEtran}
\bibliography{PGFA}

\begin{thebibliography}{10}
\providecommand{\url}[1]{#1}
\csname url@samestyle\endcsname
\providecommand{\newblock}{\relax}
\providecommand{\bibinfo}[2]{#2}
\providecommand{\BIBentrySTDinterwordspacing}{\spaceskip=0pt\relax}
\providecommand{\BIBentryALTinterwordstretchfactor}{4}
\providecommand{\BIBentryALTinterwordspacing}{\spaceskip=\fontdimen2\font plus
\BIBentryALTinterwordstretchfactor\fontdimen3\font minus \fontdimen4\font\relax}
\providecommand{\BIBforeignlanguage}[2]{{%
\expandafter\ifx\csname l@#1\endcsname\relax
\typeout{** WARNING: IEEEtran.bst: No hyphenation pattern has been}%
\typeout{** loaded for the language `#1'. Using the pattern for}%
\typeout{** the default language instead.}%
\else
\language=\csname l@#1\endcsname
\fi
#2}}
\providecommand{\BIBdecl}{\relax}
\BIBdecl

\bibitem{simonyan2014two}
K.~Simonyan and A.~Zisserman, ``Two-stream convolutional networks for action recognition in videos,'' \emph{Advances in neural information processing systems}, vol.~27, 2014.

\bibitem{tran2015learning}
D.~Tran, L.~Bourdev, R.~Fergus, L.~Torresani, and M.~Paluri, ``Learning spatiotemporal features with 3d convolutional networks,'' in \emph{Proceedings of the IEEE international conference on computer vision}, 2015, pp. 4489--4497.

\bibitem{wang2016temporal}
L.~Wang, Y.~Xiong, Z.~Wang, Y.~Qiao, D.~Lin, X.~Tang, and L.~Van~Gool, ``Temporal segment networks: Towards good practices for deep action recognition,'' in \emph{European conference on computer vision}.\hskip 1em plus 0.5em minus 0.4em\relax Springer, 2016, pp. 20--36.

\bibitem{carreira2017quo}
J.~Carreira and A.~Zisserman, ``Quo vadis, action recognition? a new model and the kinetics dataset,'' in \emph{proceedings of the IEEE Conference on Computer Vision and Pattern Recognition}, 2017, pp. 6299--6308.

\bibitem{feichtenhofer2019slowfast}
C.~Feichtenhofer, H.~Fan, J.~Malik, and K.~He, ``Slowfast networks for video recognition,'' in \emph{Proceedings of the IEEE/CVF international conference on computer vision}, 2019, pp. 6202--6211.

\bibitem{tong2022videomae}
Z.~Tong, Y.~Song, J.~Wang, and L.~Wang, ``Videomae: Masked autoencoders are data-efficient learners for self-supervised video pre-training,'' \emph{Advances in neural information processing systems}, vol.~35, pp. 10\,078--10\,093, 2022.

\bibitem{li2019deep}
C.~Li, B.~Zhang, C.~Chen, Q.~Ye, J.~Han, G.~Guo, and R.~Ji, ``Deep manifold structure transfer for action recognition,'' \emph{IEEE transactions on image processing}, vol.~28, no.~9, pp. 4646--4658, 2019.

\bibitem{liu2021semantics}
Y.~Liu, K.~Wang, G.~Li, and L.~Lin, ``Semantics-aware adaptive knowledge distillation for sensor-to-vision action recognition,'' \emph{IEEE Transactions on Image Processing}, vol.~30, pp. 5573--5588, 2021.

\bibitem{qin2015compressive}
J.~Qin, L.~Liu, Z.~Zhang, Y.~Wang, and L.~Shao, ``Compressive sequential learning for action similarity labeling,'' \emph{IEEE Transactions on Image Processing}, vol.~25, no.~2, pp. 756--769, 2015.

\bibitem{tang2019learning}
Y.~Tang, J.~Lu, Z.~Wang, M.~Yang, and J.~Zhou, ``Learning semantics-preserving attention and contextual interaction for group activity recognition,'' \emph{IEEE Transactions on Image Processing}, vol.~28, no.~10, pp. 4997--5012, 2019.

\bibitem{liu2023dual}
W.~Liu, X.~Zhong, Z.~Zhou, K.~Jiang, Z.~Wang, and C.-W. Lin, ``Dual-recommendation disentanglement network for view fuzz in action recognition,'' \emph{IEEE Transactions on Image Processing}, vol.~32, pp. 2719--2733, 2023.

\bibitem{reily2018skeleton}
B.~Reily, F.~Han, L.~E. Parker, and H.~Zhang, ``Skeleton-based bio-inspired human activity prediction for real-time human--robot interaction,'' \emph{Autonomous Robots}, vol.~42, pp. 1281--1298, 2018.

\bibitem{bandi2021skeleton}
C.~Bandi and U.~Thomas, ``Skeleton-based action recognition for human-robot interaction using self-attention mechanism,'' in \emph{2021 16th IEEE International Conference on Automatic Face and Gesture Recognition (FG 2021)}.\hskip 1em plus 0.5em minus 0.4em\relax IEEE, 2021, pp. 1--8.

\bibitem{yin2019skeleton}
J.~Yin, J.~Han, C.~Wang, B.~Zhang, and X.~Zeng, ``A skeleton-based action recognition system for medical condition detection,'' in \emph{2019 IEEE Biomedical Circuits and Systems Conference (BioCAS)}.\hskip 1em plus 0.5em minus 0.4em\relax IEEE, 2019, pp. 1--4.

\bibitem{noor2023lightweight}
N.~Noor and I.~K. Park, ``A lightweight skeleton-based 3d-cnn for real-time fall detection and action recognition,'' in \emph{Proceedings of the IEEE/CVF International Conference on Computer Vision}, 2023, pp. 2179--2188.

\bibitem{elaoud2020skeleton}
A.~Elaoud, W.~Barhoumi, E.~Zagrouba, and B.~Agrebi, ``Skeleton-based comparison of throwing motion for handball players,'' \emph{Journal of Ambient Intelligence and Humanized Computing}, vol.~11, pp. 419--431, 2020.

\bibitem{yan2018spatial}
S.~Yan, Y.~Xiong, and D.~Lin, ``Spatial temporal graph convolutional networks for skeleton-based action recognition,'' in \emph{Proceedings of the AAAI conference on artificial intelligence}, vol.~32, no.~1, 2018.

\bibitem{wang2018beyond}
H.~Wang and L.~Wang, ``Beyond joints: Learning representations from primitive geometries for skeleton-based action recognition and detection,'' \emph{IEEE Transactions on Image Processing}, vol.~27, no.~9, pp. 4382--4394, 2018.

\bibitem{shi2019two}
L.~Shi, Y.~Zhang, J.~Cheng, and H.~Lu, ``Two-stream adaptive graph convolutional networks for skeleton-based action recognition,'' in \emph{Proceedings of the IEEE/CVF conference on computer vision and pattern recognition}, 2019, pp. 12\,026--12\,035.

\bibitem{zhang2019view}
P.~Zhang, C.~Lan, J.~Xing, W.~Zeng, J.~Xue, and N.~Zheng, ``View adaptive neural networks for high performance skeleton-based human action recognition,'' \emph{IEEE transactions on pattern analysis and machine intelligence}, vol.~41, no.~8, pp. 1963--1978, 2019.

\bibitem{ye2020dynamic}
F.~Ye, S.~Pu, Q.~Zhong, C.~Li, D.~Xie, and H.~Tang, ``Dynamic gcn: Context-enriched topology learning for skeleton-based action recognition,'' in \emph{Proceedings of the 28th ACM international conference on multimedia}, 2020, pp. 55--63.

\bibitem{hao2021hypergraph}
X.~Hao, J.~Li, Y.~Guo, T.~Jiang, and M.~Yu, ``Hypergraph neural network for skeleton-based action recognition,'' \emph{IEEE Transactions on Image Processing}, vol.~30, pp. 2263--2275, 2021.

\bibitem{cheng2021extremely}
K.~Cheng, Y.~Zhang, X.~He, J.~Cheng, and H.~Lu, ``Extremely lightweight skeleton-based action recognition with shiftgcn++,'' \emph{IEEE Transactions on Image Processing}, vol.~30, pp. 7333--7348, 2021.

\bibitem{chen2021channel}
Y.~Chen, Z.~Zhang, C.~Yuan, B.~Li, Y.~Deng, and W.~Hu, ``Channel-wise topology refinement graph convolution for skeleton-based action recognition,'' in \emph{Proceedings of the IEEE/CVF international conference on computer vision}, 2021, pp. 13\,359--13\,368.

\bibitem{chi2022infogcn}
H.-g. Chi, M.~H. Ha, S.~Chi, S.~W. Lee, Q.~Huang, and K.~Ramani, ``Infogcn: Representation learning for human skeleton-based action recognition,'' in \emph{Proceedings of the IEEE/CVF conference on computer vision and pattern recognition}, 2022, pp. 20\,186--20\,196.

\bibitem{duan2022revisiting}
H.~Duan, Y.~Zhao, K.~Chen, D.~Lin, and B.~Dai, ``Revisiting skeleton-based action recognition,'' in \emph{Proceedings of the IEEE/CVF conference on computer vision and pattern recognition}, 2022, pp. 2969--2978.

\bibitem{zhu2022multilevel}
Y.~Zhu, H.~Shuai, G.~Liu, and Q.~Liu, ``Multilevel spatial--temporal excited graph network for skeleton-based action recognition,'' \emph{IEEE Transactions on Image Processing}, vol.~32, pp. 496--508, 2022.

\bibitem{myung2024degcn}
W.~Myung, N.~Su, J.-H. Xue, and G.~Wang, ``Degcn: Deformable graph convolutional networks for skeleton-based action recognition,'' \emph{IEEE Transactions on Image Processing}, vol.~33, pp. 2477--2490, 2024.

\bibitem{xu2017transductive}
X.~Xu, T.~Hospedales, and S.~Gong, ``Transductive zero-shot action recognition by word-vector embedding,'' \emph{International Journal of Computer Vision}, vol. 123, pp. 309--333, 2017.

\bibitem{jasani2019skeleton}
B.~Jasani and A.~Mazagonwalla, ``Skeleton based zero shot action recognition in joint pose-language semantic space,'' \emph{arXiv preprint arXiv:1911.11344}, 2019.

\bibitem{gupta2021syntactically}
P.~Gupta, D.~Sharma, and R.~K. Sarvadevabhatla, ``Syntactically guided generative embeddings for zero-shot skeleton action recognition,'' in \emph{2021 IEEE International Conference on Image Processing (ICIP)}.\hskip 1em plus 0.5em minus 0.4em\relax IEEE, 2021, pp. 439--443.

\bibitem{zhou2023zero}
Y.~Zhou, W.~Qiang, A.~Rao, N.~Lin, B.~Su, and J.~Wang, ``Zero-shot skeleton-based action recognition via mutual information estimation and maximization,'' in \emph{Proceedings of the 31st ACM International Conference on Multimedia}, 2023, pp. 5302--5310.

\bibitem{zhu2024part}
A.~Zhu, Q.~Ke, M.~Gong, and J.~Bailey, ``Part-aware unified representation of language and skeleton for zero-shot action recognition,'' in \emph{Proceedings of the IEEE/CVF Conference on Computer Vision and Pattern Recognition}, 2024, pp. 18\,761--18\,770.

\bibitem{he2020momentum}
K.~He, H.~Fan, Y.~Wu, S.~Xie, and R.~Girshick, ``Momentum contrast for unsupervised visual representation learning,'' in \emph{Proceedings of the IEEE/CVF conference on computer vision and pattern recognition}, 2020, pp. 9729--9738.

\bibitem{radford2021learning}
A.~Radford, J.~W. Kim, C.~Hallacy, A.~Ramesh, G.~Goh, S.~Agarwal, G.~Sastry, A.~Askell, P.~Mishkin, J.~Clark \emph{et~al.}, ``Learning transferable visual models from natural language supervision,'' in \emph{International conference on machine learning}.\hskip 1em plus 0.5em minus 0.4em\relax PMLR, 2021, pp. 8748--8763.

\bibitem{wang2023actionclip}
M.~Wang, J.~Xing, J.~Mei, Y.~Liu, and Y.~Jiang, ``Actionclip: Adapting language-image pretrained models for video action recognition,'' \emph{IEEE Transactions on Neural Networks and Learning Systems}, 2023.

\bibitem{xue2023ulip}
L.~Xue, M.~Gao, C.~Xing, R.~Mart{\'\i}n-Mart{\'\i}n, J.~Wu, C.~Xiong, R.~Xu, J.~C. Niebles, and S.~Savarese, ``Ulip: Learning a unified representation of language, images, and point clouds for 3d understanding,'' in \emph{Proceedings of the IEEE/CVF Conference on Computer Vision and Pattern Recognition}, 2023, pp. 1179--1189.

\bibitem{hegde2023clip}
D.~Hegde, J.~M.~J. Valanarasu, and V.~Patel, ``Clip goes 3d: Leveraging prompt tuning for language grounded 3d recognition,'' in \emph{Proceedings of the IEEE/CVF International Conference on Computer Vision}, 2023, pp. 2028--2038.

\bibitem{snell2017prototypical}
J.~Snell, K.~Swersky, and R.~Zemel, ``Prototypical networks for few-shot learning,'' \emph{Advances in neural information processing systems}, vol.~30, 2017.

\bibitem{huang2023location}
W.~Huang, B.~Xiao, J.~Hu, and X.~Bi, ``Location-aware transformer network for few-shot medical image segmentation,'' in \emph{2023 IEEE International Conference on Bioinformatics and Biomedicine (BIBM)}.\hskip 1em plus 0.5em minus 0.4em\relax IEEE, 2023, pp. 1150--1157.

\bibitem{rebuffi2017icarl}
S.-A. Rebuffi, A.~Kolesnikov, G.~Sperl, and C.~H. Lampert, ``icarl: Incremental classifier and representation learning,'' in \emph{Proceedings of the IEEE conference on Computer Vision and Pattern Recognition}, 2017, pp. 2001--2010.

\bibitem{lin2022prototype}
H.~Lin, Y.~Zhang, Z.~Qiu, S.~Niu, C.~Gan, Y.~Liu, and M.~Tan, ``Prototype-guided continual adaptation for class-incremental unsupervised domain adaptation,'' in \emph{European Conference on Computer Vision}.\hskip 1em plus 0.5em minus 0.4em\relax Springer, 2022, pp. 351--368.

\bibitem{iwasawa2021test}
Y.~Iwasawa and Y.~Matsuo, ``Test-time classifier adjustment module for model-agnostic domain generalization,'' \emph{Advances in Neural Information Processing Systems}, vol.~34, pp. 2427--2440, 2021.

\bibitem{du2015hierarchical}
Y.~Du, W.~Wang, and L.~Wang, ``Hierarchical recurrent neural network for skeleton based action recognition,'' in \emph{Proceedings of the IEEE conference on computer vision and pattern recognition}, 2015, pp. 1110--1118.

\bibitem{song2017end}
S.~Song, C.~Lan, J.~Xing, W.~Zeng, and J.~Liu, ``An end-to-end spatio-temporal attention model for human action recognition from skeleton data,'' in \emph{Proceedings of the AAAI conference on artificial intelligence}, vol.~31, no.~1, 2017.

\bibitem{zhang2017view}
P.~Zhang, C.~Lan, J.~Xing, W.~Zeng, J.~Xue, and N.~Zheng, ``View adaptive recurrent neural networks for high performance human action recognition from skeleton data,'' in \emph{Proceedings of the IEEE international conference on computer vision}, 2017, pp. 2117--2126.

\bibitem{choutas2018potion}
V.~Choutas, P.~Weinzaepfel, J.~Revaud, and C.~Schmid, ``Potion: Pose motion representation for action recognition,'' in \emph{Proceedings of the IEEE conference on computer vision and pattern recognition}, 2018, pp. 7024--7033.

\bibitem{caetano2019skelemotion}
C.~Caetano, J.~Sena, F.~Br{\'e}mond, J.~A. Dos~Santos, and W.~R. Schwartz, ``Skelemotion: A new representation of skeleton joint sequences based on motion information for 3d action recognition,'' in \emph{2019 16th IEEE international conference on advanced video and signal based surveillance (AVSS)}.\hskip 1em plus 0.5em minus 0.4em\relax IEEE, 2019, pp. 1--8.

\bibitem{xu2022topology}
K.~Xu, F.~Ye, Q.~Zhong, and D.~Xie, ``Topology-aware convolutional neural network for efficient skeleton-based action recognition,'' in \emph{Proceedings of the AAAI Conference on Artificial Intelligence}, vol.~36, no.~3, 2022, pp. 2866--2874.

\bibitem{cheng2020skeleton}
K.~Cheng, Y.~Zhang, X.~He, W.~Chen, J.~Cheng, and H.~Lu, ``Skeleton-based action recognition with shift graph convolutional network,'' in \emph{Proceedings of the IEEE/CVF conference on computer vision and pattern recognition}, 2020, pp. 183--192.

\bibitem{vaswani2017attention}
A.~Vaswani, N.~Shazeer, N.~Parmar, J.~Uszkoreit, L.~Jones, A.~N. Gomez, {\L}.~Kaiser, and I.~Polosukhin, ``Attention is all you need,'' \emph{Advances in neural information processing systems}, vol.~30, 2017.

\bibitem{dosovitskiy2020image}
A.~Dosovitskiy, L.~Beyer, A.~Kolesnikov, D.~Weissenborn, X.~Zhai, T.~Unterthiner, M.~Dehghani, M.~Minderer, G.~Heigold, S.~Gelly \emph{et~al.}, ``An image is worth 16x16 words: Transformers for image recognition at scale,'' \emph{arXiv preprint arXiv:2010.11929}, 2020.

\bibitem{plizzari2021spatial}
C.~Plizzari, M.~Cannici, and M.~Matteucci, ``Spatial temporal transformer network for skeleton-based action recognition,'' in \emph{Pattern recognition. ICPR international workshops and challenges: virtual event, January 10--15, 2021, Proceedings, Part III}.\hskip 1em plus 0.5em minus 0.4em\relax Springer, 2021, pp. 694--701.

\bibitem{shi2020decoupled}
L.~Shi, Y.~Zhang, J.~Cheng, and H.~Lu, ``Decoupled spatial-temporal attention network for skeleton-based action-gesture recognition,'' in \emph{Proceedings of the Asian conference on computer vision}, 2020.

\bibitem{zhang2021stst}
Y.~Zhang, B.~Wu, W.~Li, L.~Duan, and C.~Gan, ``Stst: Spatial-temporal specialized transformer for skeleton-based action recognition,'' in \emph{Proceedings of the 29th ACM International Conference on Multimedia}, 2021, pp. 3229--3237.

\bibitem{frome2013devise}
A.~Frome, G.~S. Corrado, J.~Shlens, S.~Bengio, J.~Dean, M.~Ranzato, and T.~Mikolov, ``Devise: A deep visual-semantic embedding model,'' \emph{Advances in neural information processing systems}, vol.~26, 2013.

\bibitem{sung2018learning}
F.~Sung, Y.~Yang, L.~Zhang, T.~Xiang, P.~H. Torr, and T.~M. Hospedales, ``Learning to compare: Relation network for few-shot learning,'' in \emph{Proceedings of the IEEE conference on computer vision and pattern recognition}, 2018, pp. 1199--1208.

\bibitem{word2vec}
T.~Mikolov, I.~Sutskever, K.~Chen, G.~S. Corrado, and J.~Dean, ``Distributed representations of words and phrases and their compositionality,'' in \emph{Advances in Neural Information Processing Systems 26: 27th Annual Conference on Neural Information Processing Systems 2013. Proceedings of a meeting held December 5-8, 2013, Lake Tahoe, Nevada, United States}, C.~J.~C. Burges, L.~Bottou, Z.~Ghahramani, and K.~Q. Weinberger, Eds., 2013, pp. 3111--3119.

\bibitem{reimers2019sentence}
N.~Reimers and I.~Gurevych, ``Sentence-bert: Sentence embeddings using siamese bert-networks,'' in \emph{Proceedings of the 2019 Conference on Empirical Methods in Natural Language Processing and the 9th International Joint Conference on Natural Language Processing (EMNLP-IJCNLP)}, 2019, pp. 3982--3992.

\bibitem{hubert2017learning}
Y.-H. Hubert~Tsai, L.-K. Huang, and R.~Salakhutdinov, ``Learning robust visual-semantic embeddings,'' in \emph{Proceedings of the IEEE International conference on Computer Vision}, 2017, pp. 3571--3580.

\bibitem{wray2019fine}
M.~Wray, D.~Larlus, G.~Csurka, and D.~Damen, ``Fine-grained action retrieval through multiple parts-of-speech embeddings,'' in \emph{Proceedings of the IEEE/CVF international conference on computer vision}, 2019, pp. 450--459.

\bibitem{schonfeld2019generalized}
E.~Schonfeld, S.~Ebrahimi, S.~Sinha, T.~Darrell, and Z.~Akata, ``Generalized zero-and few-shot learning via aligned variational autoencoders,'' in \emph{Proceedings of the IEEE/CVF conference on computer vision and pattern recognition}, 2019, pp. 8247--8255.

\bibitem{jia2021scaling}
C.~Jia, Y.~Yang, Y.~Xia, Y.-T. Chen, Z.~Parekh, H.~Pham, Q.~Le, Y.-H. Sung, Z.~Li, and T.~Duerig, ``Scaling up visual and vision-language representation learning with noisy text supervision,'' in \emph{International conference on machine learning}.\hskip 1em plus 0.5em minus 0.4em\relax PMLR, 2021, pp. 4904--4916.

\bibitem{wang2024cross}
X.~Wang, Y.~Yan, H.-M. Hu, B.~Li, and H.~Wang, ``Cross-modal contrastive learning network for few-shot action recognition,'' \emph{IEEE Transactions on Image Processing}, 2024.

\bibitem{zhang2022pointclip}
R.~Zhang, Z.~Guo, W.~Zhang, K.~Li, X.~Miao, B.~Cui, Y.~Qiao, P.~Gao, and H.~Li, ``Pointclip: Point cloud understanding by clip,'' in \emph{Proceedings of the IEEE/CVF conference on computer vision and pattern recognition}, 2022, pp. 8552--8562.

\bibitem{tevet2022motionclip}
G.~Tevet, B.~Gordon, A.~Hertz, A.~H. Bermano, and D.~Cohen-Or, ``Motionclip: Exposing human motion generation to clip space,'' in \emph{European Conference on Computer Vision}.\hskip 1em plus 0.5em minus 0.4em\relax Springer, 2022, pp. 358--374.

\bibitem{xiang2023generative}
W.~Xiang, C.~Li, Y.~Zhou, B.~Wang, and L.~Zhang, ``Generative action description prompts for skeleton-based action recognition,'' in \emph{Proceedings of the IEEE/CVF International Conference on Computer Vision}, 2023, pp. 10\,276--10\,285.

\bibitem{xu2022semi}
H.~Xu, L.~Liu, Q.~Bian, and Z.~Yang, ``Semi-supervised semantic segmentation with prototype-based consistency regularization,'' \emph{Advances in neural information processing systems}, vol.~35, pp. 26\,007--26\,020, 2022.

\bibitem{liu2016large}
W.~Liu, Y.~Wen, Z.~Yu, and M.~Yang, ``Large-margin softmax loss for convolutional neural networks,'' in \emph{International Conference on Machine Learning}.\hskip 1em plus 0.5em minus 0.4em\relax PMLR, 2016, pp. 507--516.

\bibitem{elsayed2018large}
G.~Elsayed, D.~Krishnan, H.~Mobahi, K.~Regan, and S.~Bengio, ``Large margin deep networks for classification,'' \emph{Advances in neural information processing systems}, vol.~31, 2018.

\bibitem{khosla2020supervised}
P.~Khosla, P.~Teterwak, C.~Wang, A.~Sarna, Y.~Tian, P.~Isola, A.~Maschinot, C.~Liu, and D.~Krishnan, ``Supervised contrastive learning,'' \emph{Advances in neural information processing systems}, vol.~33, pp. 18\,661--18\,673, 2020.

\bibitem{shi2021constrained}
Z.~Shi, H.~Wang, and C.-S. Leung, ``Constrained center loss for convolutional neural networks,'' \emph{IEEE Transactions on Neural Networks and Learning Systems}, vol.~34, no.~2, pp. 1080--1088, 2021.

\bibitem{oord2018representation}
A.~v.~d. Oord, Y.~Li, and O.~Vinyals, ``Representation learning with contrastive predictive coding,'' \emph{arXiv preprint arXiv:1807.03748}, 2018.

\bibitem{brown2020language}
T.~Brown, B.~Mann, N.~Ryder, M.~Subbiah, J.~D. Kaplan, P.~Dhariwal, A.~Neelakantan, P.~Shyam, G.~Sastry, A.~Askell \emph{et~al.}, ``Language models are few-shot learners,'' \emph{Advances in neural information processing systems}, vol.~33, pp. 1877--1901, 2020.

\bibitem{banerjee2005clustering}
A.~Banerjee, I.~S. Dhillon, J.~Ghosh, S.~Sra, and G.~Ridgeway, ``Clustering on the unit hypersphere using von mises-fisher distributions.'' \emph{Journal of Machine Learning Research}, vol.~6, no.~9, 2005.

\bibitem{wood1994simulation}
A.~T. Wood, ``Simulation of the von mises fisher distribution,'' \emph{Communications in statistics-simulation and computation}, vol.~23, no.~1, pp. 157--164, 1994.

\bibitem{shahroudy2016ntu}
A.~Shahroudy, J.~Liu, T.-T. Ng, and G.~Wang, ``Ntu rgb+ d: A large scale dataset for 3d human activity analysis,'' in \emph{Proceedings of the IEEE conference on computer vision and pattern recognition}, 2016, pp. 1010--1019.

\bibitem{liu2019ntu}
J.~Liu, A.~Shahroudy, M.~Perez, G.~Wang, L.-Y. Duan, and A.~C. Kot, ``Ntu rgb+ d 120: A large-scale benchmark for 3d human activity understanding,'' \emph{IEEE transactions on pattern analysis and machine intelligence}, vol.~42, no.~10, pp. 2684--2701, 2019.

\bibitem{liu2020benchmark}
J.~Liu, S.~Song, C.~Liu, Y.~Li, and Y.~Hu, ``A benchmark dataset and comparison study for multi-modal human action analytics,'' \emph{ACM Transactions on Multimedia Computing, Communications, and Applications (TOMM)}, vol.~16, no.~2, pp. 1--24, 2020.

\bibitem{li20213d}
L.~Li, M.~Wang, B.~Ni, H.~Wang, J.~Yang, and W.~Zhang, ``3d human action representation learning via cross-view consistency pursuit,'' in \emph{Proceedings of the IEEE/CVF conference on computer vision and pattern recognition}, 2021, pp. 4741--4750.

\bibitem{hu2024minicpm}
S.~Hu, Y.~Tu, X.~Han, C.~He, G.~Cui, X.~Long, Z.~Zheng, Y.~Fang, Y.~Huang, W.~Zhao \emph{et~al.}, ``Minicpm: Unveiling the potential of small language models with scalable training strategies,'' \emph{arXiv preprint arXiv:2404.06395}, 2024.

\bibitem{liu2024deepseek}
A.~Liu, B.~Feng, B.~Xue, B.~Wang, B.~Wu, C.~Lu, C.~Zhao, C.~Deng, C.~Zhang, C.~Ruan \emph{et~al.}, ``Deepseek-v3 technical report,'' \emph{arXiv preprint arXiv:2412.19437}, 2024.

\bibitem{ye2020few}
H.-J. Ye, H.~Hu, D.-C. Zhan, and F.~Sha, ``Few-shot learning via embedding adaptation with set-to-set functions,'' in \emph{Proceedings of the IEEE/CVF conference on computer vision and pattern recognition}, 2020, pp. 8808--8817.

\bibitem{simon2020adaptive}
C.~Simon, P.~Koniusz, R.~Nock, and M.~Harandi, ``Adaptive subspaces for few-shot learning,'' in \emph{Proceedings of the IEEE/CVF conference on computer vision and pattern recognition}, 2020, pp. 4136--4145.

\bibitem{xu2021learning}
C.~Xu, Y.~Fu, C.~Liu, C.~Wang, J.~Li, F.~Huang, L.~Zhang, and X.~Xue, ``Learning dynamic alignment via meta-filter for few-shot learning,'' in \emph{Proceedings of the IEEE/CVF conference on computer vision and pattern recognition}, 2021, pp. 5182--5191.

\bibitem{yang2024one}
S.~Yang, J.~Liu, S.~Lu, E.~M. Hwa, and A.~C. Kot, ``One-shot action recognition via multi-scale spatial-temporal skeleton matching,'' \emph{IEEE Transactions on Pattern Analysis and Machine Intelligence}, vol.~46, no.~7, pp. 5149--5156, 2024.

\bibitem{hu2024global}
J.~Hu, Y.~Hou, Z.~Guo, and J.~Gao, ``Global and local contrastive learning for self-supervised skeleton-based action recognition,'' \emph{IEEE Transactions on Circuits and Systems for Video Technology}, 2024.

\bibitem{van2008visualizing}
L.~Van~der Maaten and G.~Hinton, ``Visualizing data using t-sne.'' \emph{Journal of machine learning research}, vol.~9, no.~11, 2008.

\bibitem{mclachlan2005discriminant}
G.~J. McLachlan, \emph{Discriminant analysis and statistical pattern recognition}.\hskip 1em plus 0.5em minus 0.4em\relax John Wiley \& Sons, 2005.

\bibitem{rousseeuw1987silhouettes}
P.~J. Rousseeuw, ``Silhouettes: a graphical aid to the interpretation and validation of cluster analysis,'' \emph{Journal of computational and applied mathematics}, vol.~20, pp. 53--65, 1987.

\end{thebibliography}


\begin{thebibliography}{1}
\providecommand{\url}[1]{#1}
\csname url@samestyle\endcsname
\providecommand{\newblock}{\relax}
\providecommand{\bibinfo}[2]{#2}
\providecommand{\BIBentrySTDinterwordspacing}{\spaceskip=0pt\relax}
\providecommand{\BIBentryALTinterwordstretchfactor}{4}
\providecommand{\BIBentryALTinterwordspacing}{\spaceskip=\fontdimen2\font plus
\BIBentryALTinterwordstretchfactor\fontdimen3\font minus \fontdimen4\font\relax}
\providecommand{\BIBforeignlanguage}[2]{{%
\expandafter\ifx\csname l@#1\endcsname\relax
\typeout{** WARNING: IEEEtran.bst: No hyphenation pattern has been}%
\typeout{** loaded for the language `#1'. Using the pattern for}%
\typeout{** the default language instead.}%
\else
\language=\csname l@#1\endcsname
\fi
#2}}
\providecommand{\BIBdecl}{\relax}
\BIBdecl

\bibitem{banerjee2005clustering}
A.~Banerjee, I.~S. Dhillon, J.~Ghosh, S.~Sra, and G.~Ridgeway, ``Clustering on the unit hypersphere using von mises-fisher distributions.'' \emph{Journal of Machine Learning Research}, vol.~6, no.~9, 2005.

\bibitem{dekking2006modern}
F.~M. Dekking, C.~Kraaikamp, H.~P. Lopuha{\"a}, and L.~E. Meester, \emph{A Modern Introduction to Probability and Statistics: Understanding why and how}.\hskip 1em plus 0.5em minus 0.4em\relax Springer Science \& Business Media, 2006.

\bibitem{wood1994simulation}
A.~T. Wood, ``Simulation of the von mises fisher distribution,'' \emph{Communications in statistics-simulation and computation}, vol.~23, no.~1, pp. 157--164, 1994.

\bibitem{shahroudy2016ntu}
A.~Shahroudy, J.~Liu, T.-T. Ng, and G.~Wang, ``Ntu rgb+ d: A large scale dataset for 3d human activity analysis,'' in \emph{Proceedings of the IEEE conference on computer vision and pattern recognition}, 2016, pp. 1010--1019.

\bibitem{zhou2023zero}
Y.~Zhou, W.~Qiang, A.~Rao, N.~Lin, B.~Su, and J.~Wang, ``Zero-shot skeleton-based action recognition via mutual information estimation and maximization,'' in \emph{Proceedings of the 31st ACM International Conference on Multimedia}, 2023, pp. 5302--5310.

\bibitem{yang2024one}
S.~Yang, J.~Liu, S.~Lu, E.~M. Hwa, and A.~C. Kot, ``One-shot action recognition via multi-scale spatial-temporal skeleton matching,'' \emph{IEEE Transactions on Pattern Analysis and Machine Intelligence}, vol.~46, no.~7, pp. 5149--5156, 2024.

\end{thebibliography}
\begin{IEEEbiography}[{\includegraphics[width=1in,height=1.25in,clip,keepaspectratio]{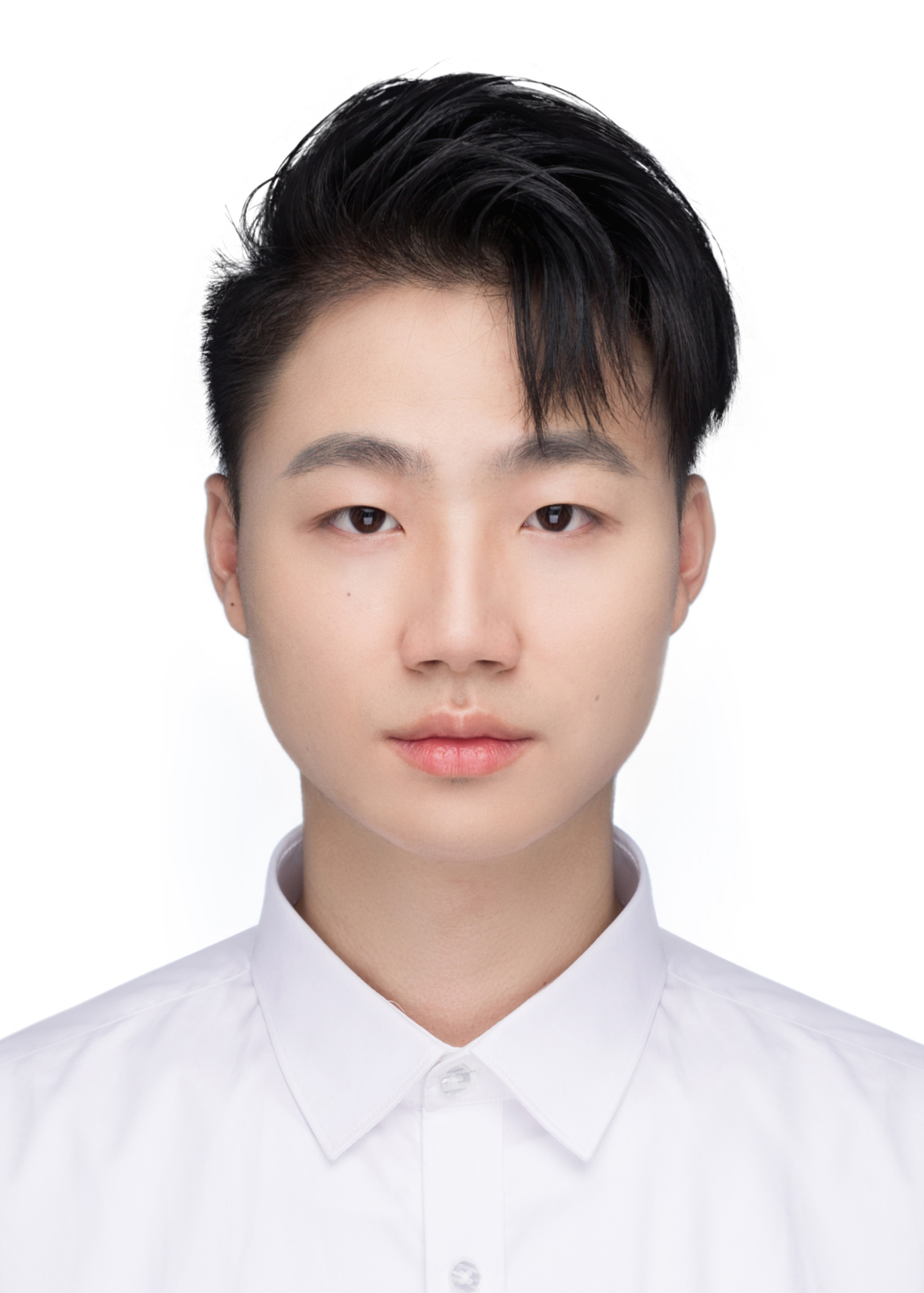}}]{Kai Zhou}
received the B.E. degree in Software Engineering from the School of Software Engineering, South China University of Technology, China, in 2020. He is currently working toward the Ph.D. degree at the same institution. His research interests include deep learning and computer vision.
\end{IEEEbiography}

\begin{IEEEbiography}[{\includegraphics[width=1in,height=1.25in,clip,keepaspectratio]{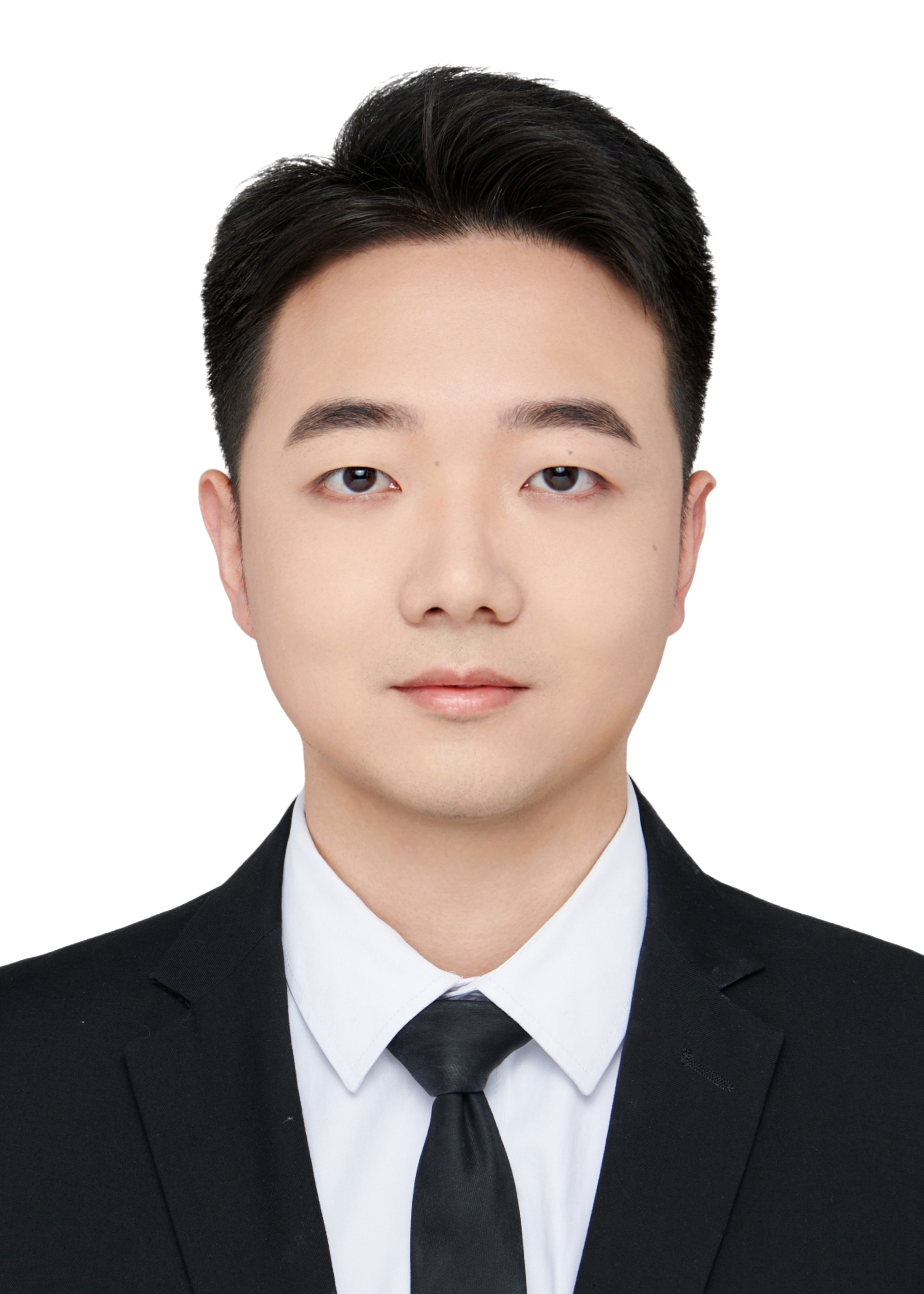}}]{Shuhai Zhang}
is currently a Ph.D. candidate at South China University of Technology, China. His research interests are broadly in machine learning and mainly focus on large language model, model compression, and adversarial robust. He has published papers in Neural Networks, T-CSVT, ICCV, ICML, ICLR, CVPR.
\end{IEEEbiography}

\begin{IEEEbiography}[{\includegraphics[width=1in,height=1.25in,clip,keepaspectratio]{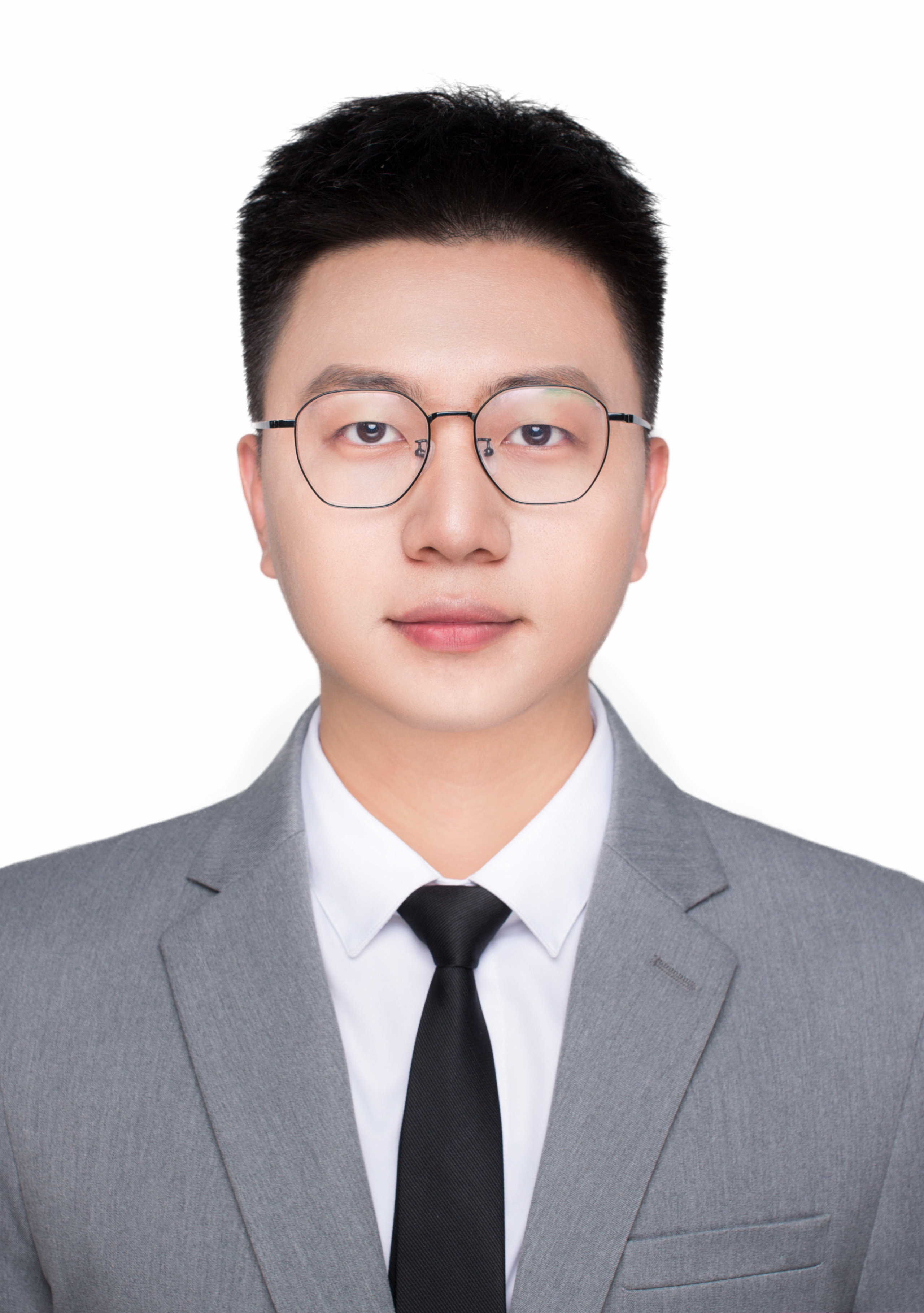}}]{Zeng You}
received the B.E. degree in Software Engineering from the School of Software Engineering, South China University of Technology, China, in 2020. He is currently working toward the Ph.D. degree in the School of Future Technology, South China University of Technology, China. His research interests include deep learning, video understanding, and computer vision.
\end{IEEEbiography}

\begin{IEEEbiography}[{\includegraphics[width=0.9in,height=1.25in, clip,keepaspectratio]{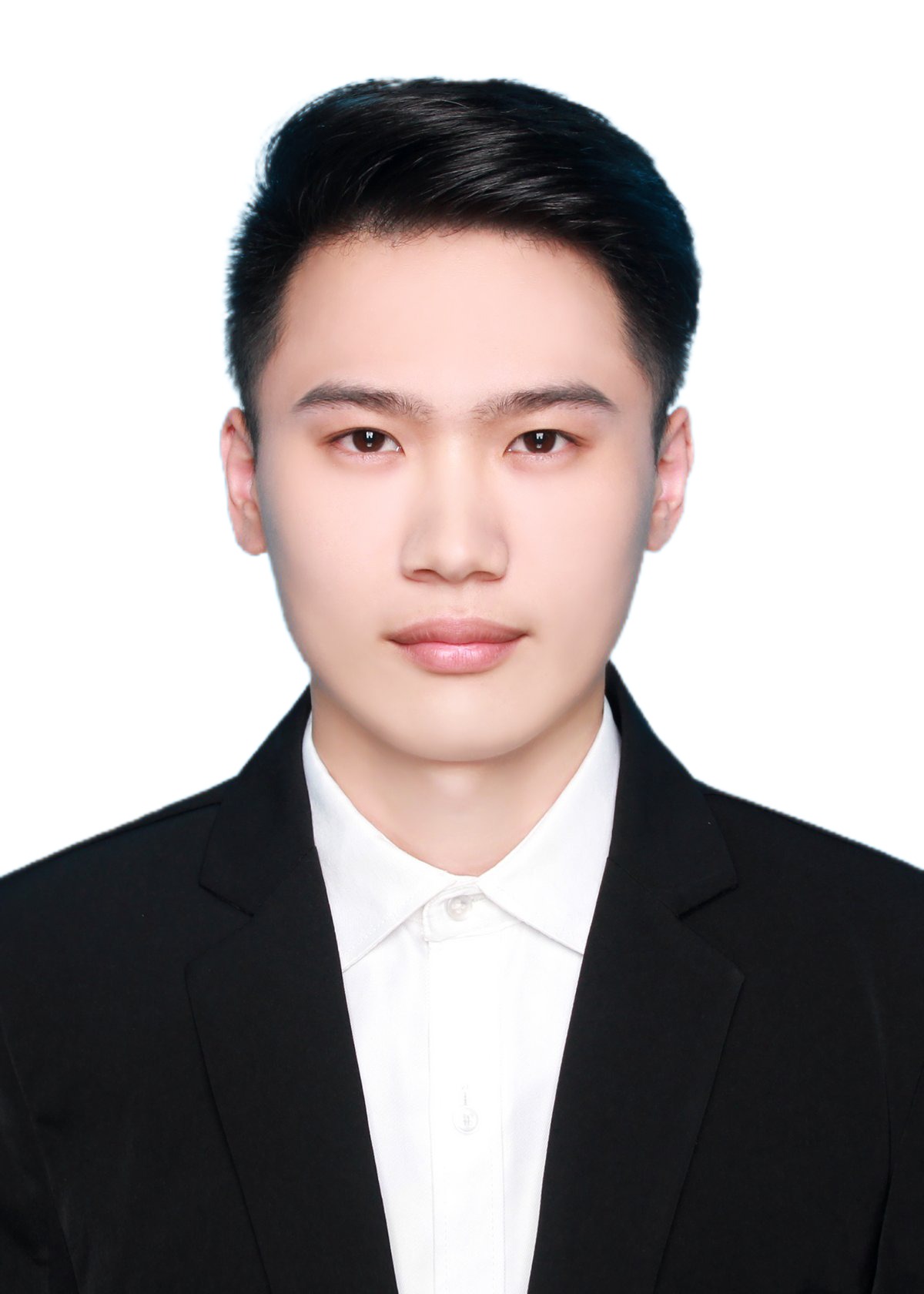}}]{Jinwu Hu} (Graduate Student Member, IEEE) received the B.E. degree from Foshan University, Foshan, China, in 2020, and the M.S. degree from the Chongqing University of Posts and Telecommunications, Chongqing, China, in 2023. He is currently pursuing a Ph.D. degree at the South China University of Technology, Guangzhou,  China. His research interests include computer vision, machine learning, large language models, and reinforcement learning. He has published several journal/conference papers, including IEEE TCYB, IEEE TMI, IEEE TBD, ICML, IJCAI, CVPR, ACM MM, etc. He has served as a reviewer for many academic journals, including IEEE JBHI, NN, PR, etc.
\end{IEEEbiography}

\begin{IEEEbiography}[{\includegraphics[width=1in,height=1.25in,clip,keepaspectratio]{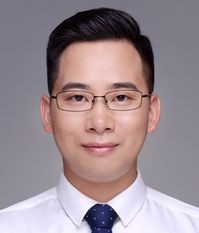}}]{Mingkui Tan} (Senior Member, IEEE)
is currently a professor with the School of Software Engineering at South China University of Technology. He received his Bachelor's Degree in Environmental Science and Engineering in 2006 and Master's degree in Control Science and Engineering in 2009, both from Hunan University in Changsha, China. He received his Ph.D. degree in Computer Science from Nanyang Technological University, Singapore, in 2014. From 2014-2016, he worked as a Senior Research Associate on computer vision in the School of Computer Science, University of Adelaide, Australia. His research interests include machine learning, sparse analysis, deep learning, and large-scale optimization.
\end{IEEEbiography}

\begin{IEEEbiography}[{\includegraphics[width=1in,height=1.25in,clip,keepaspectratio]{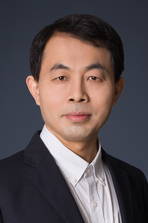}}]{Fei Liu} (Member, IEEE) received the B.E. degree in automatic testing and control, M.E. and Ph.D. degrees in control science and engineering from Harbin Institute of Technology. He worked at department of computer science in Brandenburg University of Technology from 2009 to 2012 and at department of computer in Yamaguchi University from 2015 to 2016 as a JSPS scholar. He is currently a professor in the School of Software Engineering, South China University of Technology. His research interests include modeling and simulation, artificial intelligence, and systems biology.
\end{IEEEbiography}

\vfill

\end{document}

\title{Supplementary Materials for \\``Zero-Shot Skeleton-Based Action Recognition With Prototype-Guided Feature Alignment"}
\author{Kai Zhou, Shuhai Zhang, Zeng You, Jinwu Hu, Mingkui Tan, \IEEEmembership{Senior Member, IEEE}, \\and Fei Liu, \IEEEmembership{Member, IEEE}
        % <-this % stops a space

\thanks{This work was supported by National Natural Science Foundation of China (62072190, U24A20327, U23B2013, and 62276176), and 
Guangdong Basic and Applied Basic Research Foundation (2024A1515010900). \textit{(Kai Zhou and Shuhai Zhang contributed equally to this work.) (Corresponding authors: Mingkui Tan; Fei Liu.)}}

\thanks{Kai Zhou and Fei Liu are with the School of Software Engineering, South China University of Technology, Guangzhou, China (e-mail: kayjoe0723@gmail.com, feiliu@scut.edu.cn).}

\thanks{Shuhai Zhang and Jinwu Hu are with the School of Software Engineering, South China University of Technology, and with Pazhou Lab, Guangzhou, China (e-mail: shuhaizhangshz@gmail.com, fhujinwu@gmail.com).}

\thanks{Zeng You is with the School of Future Technology, South China University of Technology, Guangzhou, China and also with Peng Cheng Laboratory, Shenzhen, China (e-mail: zengyou.yz@gmail.com).}

\thanks{Mingkui Tan is with the School of Software Engineering, South China University of Technology, Guangzhou, China, and also with the Key Laboratory of Big Data and Intelligent Robot (South China University of Technology), Ministry of Education, Guangzhou 510006, China (e-mail: mingkuitan@scut.edu.cn).}
}% <-this % stops a space
\markboth{IEEE TRANSACTIONS ON IMAGE PROCESSING}% , VOL. ~, NO. ~, August~2021
{Zhou \MakeLowercase{\textit{et al.}}: Zero-shot Skeleton-based Action Recognition With Prototype-Guided Feature Alignment}


\maketitle

\nolinenumbers

We organize the supplementary into the following sections. In Section \ref{sec:thm}, we derive the proof of the theorem \ref{thm: argP}. In Section \ref{sec:vis}, we report more qualitative analysis of our \sexyname. \major{In Section \ref{sec:one-shot}, we provide the detailed dataset splits of our one-shot experiments.}

\section{Proof of Theorem 1}
\label{sec:thm}
\begin{thm}
\label{thm: argP}
    Assuming that the distributions of the $K$ classes of normalized skeleton feature $\bv^{(k)}$ are from $K$ von Mises–Fisher distributions \cite{banerjee2005clustering} with the same concentration parameter $\kappa$ but different mean directions $\boldsymbol{\mu}_k$, i.e., $f_{p_k}(\bv;\boldsymbol{\mu}_k,\kappa)=C_{d}(\kappa)\exp(\kappa\boldsymbol{\mu}_k^\mathsf{T}\mathbf{v})$, $k=1, \ldots, K$, where $C_{d}(\kappa)$ is a normalization constant related to the concentration parameter $\kappa$ and feature dimension $d$,
    and the $k$-th class of prototype feature is $\hat{\boldsymbol{\mu}}_k= \frac{\frac{1}{n}\sum_{i=1}^n \bv_i^{(k)}}{\|\frac{1}{n}\sum_{i=1}^n \bv_i^{(k)}\|} $, which is a normalized $\bc^{u,k}$.
    Then, as $n \to \infty$, for $\forall~\bv \sim p_m$, $m=1,\ldots K$, if $k=\mathop{\arg\max}\limits_{i}\frac{\exp(\mathrm{sim}(\bv,\hat{\boldsymbol{\mu}}_i))}{\sum_{j=1}^{K}\exp(\mathrm{sim}(\bv,\hat{\boldsymbol{\mu}}_j))}$, we have
    \begin{equation}
        P(\bv | \mathrm{class}~k)>P(\bv | \mathrm{class}~j), \forall~j \neq k.
    \end{equation}
\end{thm}

\begin{proof}
    According to the $k=\mathop{\arg\max}\limits_{i}\frac{\exp(\mathrm{sim}(\bv,\hat{\boldsymbol{\mu}}_i))}{\sum_{j=1}^{K}\exp(\mathrm{sim}(\bv,\hat{\boldsymbol{\mu}}_j))}$, we have $k=\mathop{\arg\max}\limits_{i} \mathrm{sim}(\bv,\hat{\boldsymbol{\mu}}_i)=\mathop{\arg\max}\limits_{i}  \hat{\boldsymbol{\mu}}_i^\top \bv$ with $\|\bv\|=\|\hat{\boldsymbol{\mu}}_i\|=1$. This means 
    \begin{equation}
    \label{neq: hat}
        \hat{\boldsymbol{\mu}}_k^\top \bv> \hat{\boldsymbol{\mu}}_j^\top \bv, \forall~j \neq k.
    \end{equation}
    Based on the law of large numbers \cite{dekking2006modern}, we have 
    \begin{equation}
    \label{eq: lim}
        \lim \limits_{n \to \infty} \hat{\boldsymbol{\mu}}_k^\top \bv {=} \lim \limits_{n \to \infty} \frac{\left(\frac{1}{n}\sum_{i=1}^n \bv_i^{(k)}\right)^\top \bv}{\|\frac{1}{n}\sum_{i=1}^n \bv_i^{(k)}\|}{=} \frac{{A_d(\kappa)\boldsymbol{\mu}}_k^\top \bv}{A_d(\kappa)}{=}\boldsymbol{\mu}_k^\top \bv,
    \end{equation}
    where $A_d(\kappa)$ is a constant related to the concentration parameter $\kappa$ and dimension $d$ \cite{banerjee2005clustering, wood1994simulation}.
    The limit of the numerator is based on the property of von Mises-Fisher distribution, while the denominator is based on 
    \begin{align}
       ~~&\lim \limits_{n \to \infty}
       \left|\|\frac{1}{n}\sum_{i=1}^n \bv_i^{(k)}\| - A_d(\kappa) \|\boldsymbol{\mu}_k \| \right| \\
       {\leq}& 
       \lim \limits_{n \to \infty}
       \left\| \frac{1}{n}\sum_{i=1}^n \bv_i^{(k)} - A_d(\kappa) \boldsymbol{\mu}_k  \right\|=0,
    \end{align}
    which leads to
    \begin{align}
        \lim \limits_{n \to \infty} \left \|\frac{1}{n}\sum_{i=1}^n \bv_i^{(k)} \right\|= A_d(\kappa)\|\boldsymbol{\mu}_k\|=A_d(\kappa).
    \end{align}
    Combing Eq. (\ref{neq: hat}) and Eq. (\ref{eq: lim}), as $n \to \infty$, we have 
        \begin{equation}
        {\boldsymbol{\mu}}_k^\top \bv> {\boldsymbol{\mu}}_j^\top \bv, \forall~j \neq k.
    \end{equation}
    Thus, as $n \to \infty$, we get the result:
    \begin{equation}
        C_{d}(\kappa)\exp(\kappa\boldsymbol{\mu}_k^\mathsf{T}\mathbf{v}) > C_{d}(\kappa)\exp(\kappa\boldsymbol{\mu}_j^\mathsf{T}\mathbf{v}), \forall~j \neq k,
    \end{equation}
    Therefore, we obtain the conclusion:
        \begin{equation}
        P(\bv | \mathrm{class}~k)>P(\bv | \mathrm{class}~j), \forall~j \neq k.
    \end{equation}
\end{proof}

\section{More Qualitative Analysis}
\label{sec:vis}
\begin{figure*}[tb]
    \centering
    \includegraphics[width=1.0\textwidth]{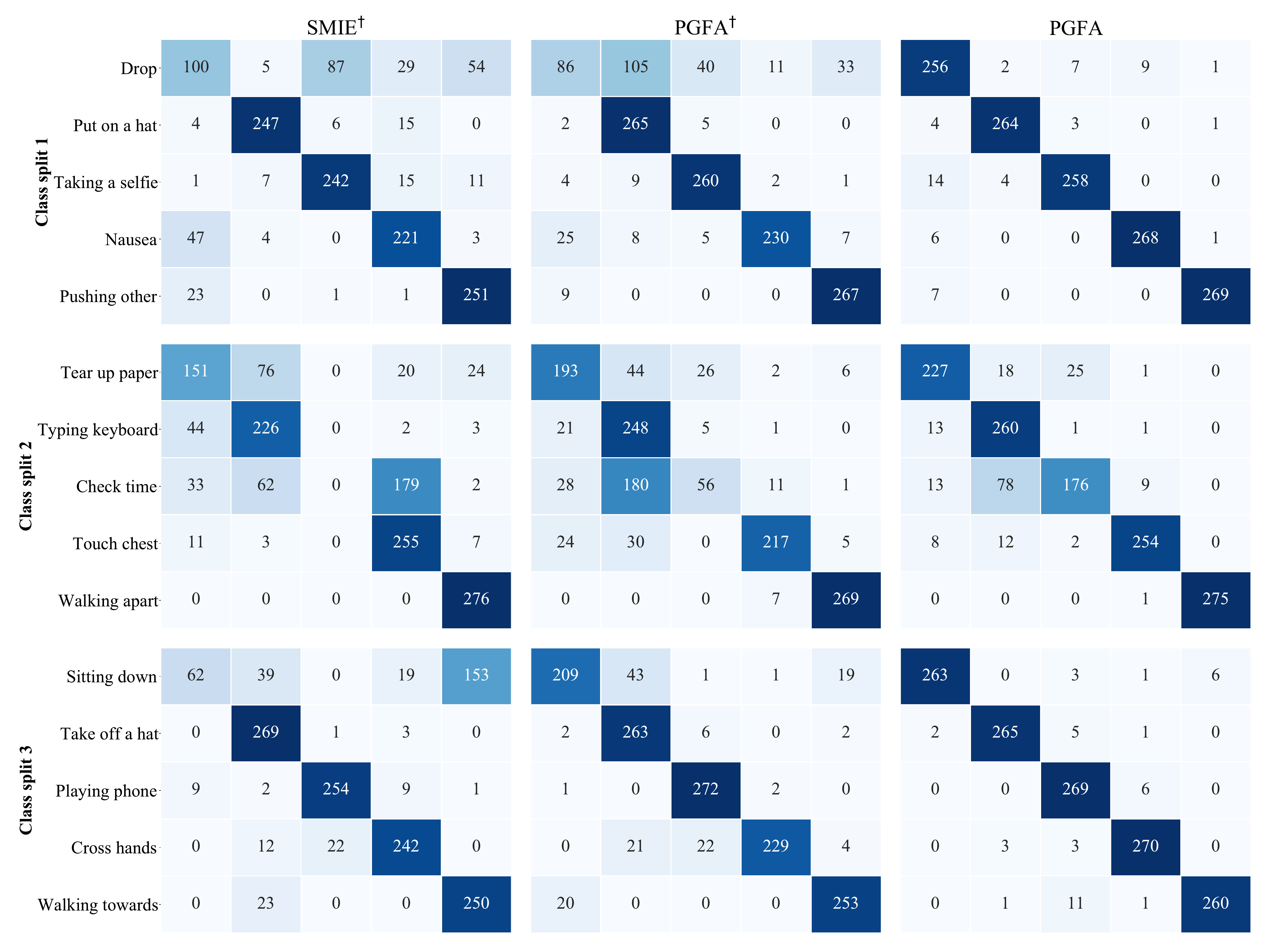}
    \caption{The confusion matrix visualization for NTU-60 in Setting \uppercase\expandafter{\romannumeral1} compares three methods: SMIE (reimplemented) in the first column, our \sexynameplus$^\dag$ (without prototype-guided text feature alignment) in the second column, and our full method \sexyname in the third column. The x-axis of each confusion matrix represents the predicted unseen class, while the y-axis represents the true class. The labels on the y-axis represent the class names of the unseen actions from three different class splits.}
    \label{fig:heatmap}
\end{figure*}

\textbf{Visualization of the confusion matrices.} To more intuitively observe each category’s classification accuracy, we visualize confusion matrices for the three 55/5 (seen/unseen) class splits on the NTU-60 \cite{shahroudy2016ntu} dataset (Setting \uppercase\expandafter{\romannumeral1}). In Fig. \ref{fig:heatmap}, we present the confusion matrix visualizations for SMIE \cite{zhou2023zero} and our method. All models are trained using complete descriptions as text input. Each confusion matrix includes five unseen classes, and the value at each position represents the number of samples classified into that class. From Fig. \ref{fig:heatmap}, we observe that the SMIE method is prone to misclassify hand movement actions, such as ``Check time" and ``Touch chest" in Class Split 2. Our \sexynameplus$^\dag$ partially improved this issue, but misclassification still occurs, such as with ``Check time" and ``Typing keyboard". In contrast, our full method \sexyname, through the prototype-guided text feature alignment strategy, significantly improved the accuracy of the ``Check time" action. This demonstrates the necessity of aligning skeleton and unseen text features during the testing phase.

\section{Dataset splitting of one-shot experiments}\label{sec:one-shot}
\major{Following \cite{yang2024one}, we conduct one-shot skeleton-based action recognition experiments on the NTU-60, NTU-120, and PKU-MMD datasets. The seen/unseen class splits for each dataset are as follows:}

\major{
\textbf{(1) NTU-60.} We select 10 classes as unseen classes and 10 corresponding exemplars for reference during testing, while the remaining 50 seen classes are used for training. These 10 unseen classes are: A1 (drink water), A7 (throw), A13 (tear up paper), A19 (take off glasses), A25 (reach into pocket), A31 (pointing to something with finger), A37 (wipe face), A43 (falling), A49 (use a fan (with hand or paper)/feeling warm), A55 (hugging another person). The 10 corresponding exemplars are: 'S001C003P008R001A001', 'S001C003P008R001A007', 'S001C003P008R001A013', 'S001C003P008R001A019', 'S001C003P008R001A025', 'S001C003P008R001A031', 'S001C003P008R001A037', 'S001C003P008R001A043', 'S001C003P008R001A049', 'S001C003P008R001A055'.}

\major{
\textbf{(2) NTU-120.} We select 20 classes as unseen classes and 20 corresponding exemplars for reference during testing, while the remaining 100 classes are used for training. These 20 unseen classes are: A1 (drink water), A7 (throw), A13 (tear up paper), A19 (take off glasses), A25 (reach into pocket), A31 (pointing to something with finger), A37 (wipe face), A43 (falling), A49 (use a fan (with hand or paper)/feeling warm), A55 (hugging another person), A61 (put on headphones), A67 (hush/quiet), A73 (staple book), A79 (sniff/smell), A85 (apply cream on face), A91 (open a box), A97 (arm circles), A103 (yawn), A109 (grab other person's stuff), A115 (take a photo of another person). The 20 corresponding exemplars are:
'S001C003P008R001A001',
'S001C003P008R001A007',
'S001C003P008R001A013',
'S001C003P008R001A019',
'S001C003P008R001A025',
'S001C003P008R001A031',
'S001C003P008R001A037',
'S018C003P008R001A043',
'S018C003P008R001A049',
'S018C003P008R001A055',
'S018C003P008R001A061',
'S018C003P008R001A067',
'S018C003P008R001A073',
'S018C003P008R001A079',
'S001C003P008R001A085',
'S001C003P008R001A091',
'S001C003P008R001A097',
'S018C003P008R001A103',
'S018C003P008R001A109',
'S018C003P008R001A115'.}

\major{\textbf{(3) PKU-MMD.} We select 10 classes as unseen classes and 10 corresponding exemplars for reference during testing, while the remaining 41 classes are used for training. These 10 unseen classes are: A1 (bow), A6 (clapping), A11 (falling), A16 (hugging another person), A21 (pat on the back of another person), A26 (punching/slapping other person), A31 (rub two hands together), A36 (take off glasses), A41 (throw), A46 (typing on a keyboard). The 10 corresponding exemplars are: 'F003V001C001', 'F003V001C006', 'F002V001C011', 'F005V001C016', 'F005V001C021', 'F005V001C026', 'F002V001C031', 'F003V001C036', 'F002V001C041', 'F003V001C046'. Note that the exemplar IDs above slightly differ from the official ones. For example, 'F003V001C001' corresponds to the skeleton sequence segment of video '0003' (F003), the left-side view (V001), and action category 1 (C001).}

\bibliographystyle{IEEEtran}
\bibliography{PGFA}

\vfill